\documentclass[journal]{IEEEtran}
\usepackage{array}
\usepackage{cite}
\usepackage{amsmath,graphicx}
\usepackage{epstopdf}
\usepackage{xcolor}
\usepackage{amssymb}
\usepackage{pifont}
\usepackage{colortbl, booktabs}
\usepackage{booktabs}
\usepackage{array}
\usepackage{tabu}
\usepackage{booktabs}
\usepackage{subfigure}
\usepackage{pifont}
\usepackage{bbding}
\usepackage{verbatim}
\usepackage{ulem}
\usepackage{multirow}
\usepackage{float}
\usepackage{stfloats}
\usepackage[misc]{ifsym}
\usepackage{stmaryrd}
\usepackage{subfigure}
\usepackage{grffile}
\usepackage{makecell}
\usepackage{threeparttable}
\usepackage{bm}
\usepackage{enumitem}
\usepackage{soul}
\usepackage{color} 
\usepackage{enumerate}
\usepackage[pagebackref=false,breaklinks=true,colorlinks,bookmarks=false]{hyperref}

\begin{document}
\title{DuInNet: Dual-Modality Feature Interaction for Point Cloud Completion}

\author{Xinpu Liu, Baolin Hou, Hanyun Wang, Ke Xu, Jianwei Wan, Yulan Guo,~\IEEEmembership{Senior Member,~IEEE}
        % <-this % stops a space

\thanks{This work was supported in part by the National Natural Science Foundation of China (No. 42271457) and the National Key Research and Development Program of China (No. 2022YFB3902400). \textit{(Corresponding author: Hanyun Wang)}}

\thanks{Xinpu Liu, Baolin Hou, Ke Xu, Jianwei Wan and Yulan Guo are with the College of Electronic Science and Technology, National University of Defense Technology (NUDT), Changsha 410000, China. Emails: \{liuxinpu, houbaolin, xuke, wanjianwei, yulan.guo\}@nudt.edu.cn. \textit{(Xinpu Liu and Baolin Hou contribute equally.)}}

\thanks{Hanyun Wang is with the School of Surveying and Mapping, Information Engineering University, Zhengzhou 450001, China. Email: why.scholar@gmail.com.}}

\markboth{Journal of \LaTeX\ Class Files,~Vol.~14, No.~8, August~2021}%
{Shell \MakeLowercase{\textit{et al.}}: A Sample Article Using IEEEtran.cls for IEEE Journals}


\maketitle

\begin{abstract}
To further promote the development of multimodal point cloud completion, we contribute a large-scale multimodal point cloud completion benchmark ModelNet-MPC with richer shape categories and more diverse test data, {which contains nearly 400,000 pairs of high-quality point clouds and rendered images of 40 categories.}
Besides the fully supervised point cloud completion task, two additional tasks including denoising {completion} and zero-shot learning {completion} are proposed in ModelNet-MPC, to {simulate real-world scenarios} and verify the robustness to noise and the transfer ability across categories of current methods. 
Meanwhile, considering that existing multimodal completion pipelines usually adopt a unidirectional fusion mechanism and ignore the shape prior contained in the image modality, we propose a Dual-Modality Feature Interaction Network (DuInNet) in this paper.
DuInNet iteratively interacts features between point clouds and images to learn both geometric and texture characteristics of shapes with the dual feature interactor.
To adapt to specific tasks such as fully supervised, denoising, and zero-shot learning {point cloud completions}, an adaptive point generator is proposed to generate complete point clouds in blocks with different weights for these two modalities. 
Extensive experiments on the ShapeNet-ViPC and ModelNet-MPC benchmarks demonstrate that DuInNet exhibits superiority, robustness and transfer ability {in all completion tasks} over state-of-the-art methods.
The code and dataset will be available soon.
\end{abstract}


\section{Introduction}
\label{sec:intro}

\IEEEPARstart{P}{oint} cloud is commonly used to represent the geometric information of 3D shapes\cite{review}.
However, due to various limits such as sensor resolution, light reflection, self-occlusion and viewing angles, the point clouds captured by existing sensors are usually incomplete and noisy, posing challenges to downstream computer vision tasks, such as point cloud classification\cite{pointbench}, part segmentation\cite{pointnext}, object detection\cite{iassd}, and semantic segmentation\cite{2dpass}.
Therefore, recovering complete point clouds of 3D shapes from partial observations is essential for practical applications.

\begin{figure}[t]
  \centering
  \subfigure[Single]{
  \begin{minipage}[t]{0.19\linewidth}
    \centering
    \includegraphics[width=1\linewidth]{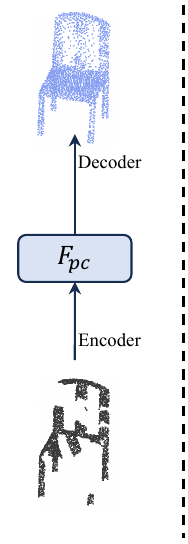}
    \end{minipage}
    \label{fig:overview-a}
  }
  \subfigure[Unidirectional fusion]{
  \begin{minipage}[t]{0.355\linewidth}
    \centering
    \includegraphics[width=1\linewidth]{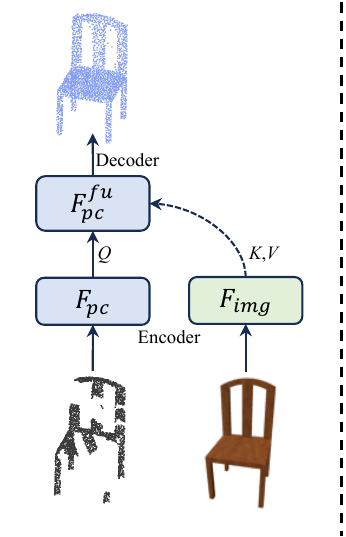}
    \end{minipage}
    \label{fig:overview-b}
  }
  \subfigure[Dual interaction]{
  \begin{minipage}[t]{0.31\linewidth}
    \centering
    \includegraphics[width=1\linewidth]{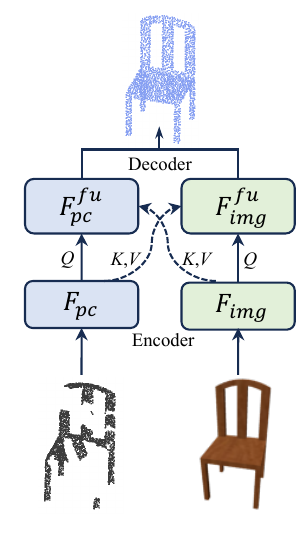}
    \end{minipage}
    \label{fig:overview-c}
  }
   \caption{\textbf{Schematic completion strategy comparison.} \textbf{(a)} Restoring complete shapes through autoencoders of partial point clouds. \textbf{(b)} Images-assisted point cloud completion by \textbf{unidirectional} fusion. \textbf{(c)} Our image and point cloud \textbf{dual} interaction strategy for shapes completion. Here, $F$ represents a feature matrix, $Q$, $K$ and $V$ represent query, key and value tensors, respectively.} 
   \label{fig:overview}
\end{figure}
{Completing partial point clouds to restore their full shapes is an ill-posed and challenging problem, as it requires a amount of shape prior information.}
According to the input modalities, current point cloud completion methods can be divided into two categories: single-modal and multimodal methods.
As shown in Fig.~\ref{fig:overview-a}, most single-modal point cloud completion methods are based on the structure of autoencoders\cite{pfnet,vrcnet,ecg,foldingnet,atlasnet,sdt,msn,pcn,topnet,seedformer}, which reconstruct complete point clouds by decoding feature vectors. However, these methods are sensitive to the quality of partial point clouds.

Images are more natural to human vision and can provide additional shape priors, such as rich texture and clear object boundaries\cite{swin,deepinteraction,pointcmt}. Adding images into the point cloud completion task pipeline can help to infer the shapes of missing parts based on their 2D experience.
Current multimodal methods primarily take 3D point clouds and 2D images as their input, and combine these individual modality features into a unidirectional hybrid representation\cite{vipc,csdn,xmfnet}, as shown in Fig.~\ref{fig:overview-b}. 
{
The pioneer work ViPC\cite{vipc} explicitly reconstructs coarse point clouds from two modalities separately. However, generating a coarse point cloud from a single image is a challenging problem and susceptible to noises. 
Although CSDN\cite{csdn} avoids the explicit process of reconstructing point clouds from images, it fails to disengage from the pattern of feature concatenation and has limited ability to distinguish shape prior and noise, resulting in poor performance of point cloud completion. 
}
XMFNet\cite{xmfnet} utilizes the cross-attention mechanism\cite{attention} to interact features between two modalities, and generates complete point clouds in blocks.
However, such a single hybrid representation primarily enhances the features of partial point clouds, but fails to explicitly highlight benefits from image features. Consequently, its generated point clouds are blurry and noisy.

In this paper, we propose a Dual-Modality Feature Interaction Network (DuInNet) for point cloud completion (Fig.~\ref{fig:overview-c}), which consists of two encoders for two modalities, a dual feature interactor (DFI), and an adaptive point generator (APG).
{After extracting features from point clouds and images separately,} DFI aims to iteratively facilitate feature interaction between point clouds and images, enabling the network to learn both geometric and texture characteristics of shapes.
Specifically, DFI performs features interaction in a latent domain via a dual-path architecture, which alternately sets images or point clouds as a query matrix to retrieve and fuse information from the shape features of the other modality.
APG is a flexible decoder which aims to ensure that these two fused features both contribute to the completion process. 
Specifically, APG utilizes multi-layer perceptrons to perform shape mapping based on the fusion features generated by DFI, and generates point clouds with different weights for two different modalities.

Currently, there is only one multimodal dataset ShapeNet-ViPC\cite{vipc} for point cloud completion. ShapeNet-ViPC contains point clouds and their corresponding images of 13 categories. However, the diversity of shape categories, the robustness to noise, and the transfer ability to unseen categories of point cloud completion methods cannot be fully explored with the ShapeNet-ViPC.
Therefore, we create a new large-scale ModelNet-based multimodal point cloud completion dataset (entitled as ModelNet-MPC) with nearly 400,000 pairs of high-quality point clouds and {rendered} images for 40 categories.
To improve data diversity, for each complete 3D CAD model in ModelNet40\cite{modelnet}, we first select 32 viewpoints uniformly distributed on a unit sphere, then render partial point clouds and their corresponding images for each viewpoint.
Meanwhile, to demonstrate the robustness to noise and the transfer ability across categories of existing {completion} methods, we also set up two new tasks (\textit{i.e.} denoising {completion} and zero-shot learning {completion}) for our ModelNet-MPC dataset, besides the fully supervised multimodal point cloud completion task.

Our contributions are summarized as follows:
\begin{itemize}[leftmargin=2em]
\item We propose DuInNet to restore complete point clouds with the shape priors from both images and point clouds. In addition, DuInNet can be adapted to different point cloud completion tasks.
\item We propose a new large-scale multimodal point cloud completion dataset ModelNet-MPC, which provides a comprehensive benchmark for the evaluation of point cloud completion methods, including the robustness to noise, generalization and transfer ability across different categories.
\item Extensive experiments on the ShapeNet-ViPC and ModelNet-MPC datasets demonstrate the superiority and generalization ability of DuInNet. In particular, denoising and zero-shot learning {completion} experiments demonstrate the robustness and transfer ability {to unseen categories} of DuInNet.
\end{itemize}

{
This paper is organized as follows: 
In Section \ref{sec:related}, we briefly review the related work. 
In Section \ref{sec:DuInNet}, we introduce the architecture of our DuInNet in details.
In Section \ref{sec:ModelNet-MPC}, we introduce our self-developed ModelNet-MPC dataset in details. 
The experimental results are presented in Section \ref{sec:experiments}.
Section \ref{sec:conclusion} gives the conclusion.
}
\section{Related Work}
\label{sec:related}

Point cloud completion is an important task in computer vision and graphics.
Early methods mainly adopt shape retrieval or template feature matching to generate complete point clouds~\cite{ct1,ct2,ct3}. However, these methods heavily rely on the quality of the database, resulting in poor performance and scalability.
With the development of deep neural networks for 3D point clouds~\cite{pn,pn++,dgcnn,pt,paconv}, numerous classical network architectures have been developed for point cloud completion.
In this section, we first briefly review the major learning-based methods on single-modal and multi-modal point cloud completion, and then introduce the corresponding datasets, {and finally, we briefly review the multi-modal feature interaction technology.
}

\subsection{Single-modal point cloud completion}
PCN~\cite{pcn} first introduces an autoencoder structure into the point cloud generation process, which learns global features from partial point clouds and restores complete ones with a folding method~\cite{foldingnet}.
TopNet~\cite{topnet} proposes a novel decoder with a hierarchical tree structure, enabling the generation of structurally refined point clouds without assuming any specific structure or topology.
GRNet~\cite{grnet} introduces 3D grids as intermediate representations to regularize unordered point clouds, and uses 3D convolution to extract features and generate points.
Additionally, PoinTr~\cite{pointr} considers point cloud completion as a set-to-set translation problem and first proposes a fully transformer-based completion network. 
SnowflakeNet~\cite{snowflakenet} utilizes self-attentions on points to extract discriminative features, and adopts the skip-transformer to split parent points and gradually generate child points.
{AGFA-Net~\cite{AGFA} utilizes spatial attention blocks to replace KNN operations and aggregate global features adaptively by calculating per-point attention values for point cloud generation.}
To enhance the ability to preserve and recover details of 3D shapes, SeedFormer~\cite{seedformer} introduces patch seeds as a new shape representation, and integrates seed features into the completion process to generate faithful details of point clouds in a coarse-to-fine manner. 
Although these methods have achieved good performances in point cloud completion, it is still challenging to generate high-precision point clouds with single-modality if a large part of points are missing. 

\subsection{Multi-modal point cloud completion}
Recently, point cloud completion with multi-modalities has attracted increasing attention.
ViPC~\cite{vipc} utilizes additional images to search for missing global structural features to generate complete point clouds. Meanwhile, ViPC proposes a multimodal dataset (namely, ShapeNet-ViPC) based on the existing ShapeNet dataset~\cite{shapenet}.
To simulate the physical repair process of point cloud completion, CSDN~\cite{csdn} utilizes shape fusion and dual refinement modules to generate complete shapes. It not only explores complementary information from images but also effectively leverages cross-modal data throughout a coarse-to-fine completion process. 
Furthermore, XMFNet~\cite{xmfnet} effectively combines features from two modalities using a cross-attention mechanism and avoids the need for point cloud reconstruction solely from images. Additionally, XMFNet also proposes a weakly-supervised method to achieve image-assistant point cloud shape completion. 
However, these methods primarily enhance features of point clouds through a single hybrid representation, but fail to fully explore the benefits of images.

\subsection{Datasets for point cloud completion}
The PCN~\cite{pcn} dataset is a widely used point cloud completion benchmark derived from the ShapeNet~\cite{shapenet} dataset, and it contains 30,974 complete-partial point cloud pairs of 8 categories.
In PCN, the partial point clouds are generated by back-projecting complete 3D shapes of ShapeNet dataset into 8 different views, and each ground-truth point cloud has 16,384 points, which are evenly sampled from the shape surface.
The Completion3D~\cite{topnet} dataset is also a subset of the ShapeNet~\cite{shapenet} dataset containing 30,958 models with 8 categories, and both partial and ground truth point clouds have 2048 points.
The MVP~\cite{vrcnet} dataset is a high-quality multi-view point cloud completion benchmark. It has 16 categories and 26 virtual views, and consists of point clouds with diverse scales and resolutions.
The ShapeNet-55~\cite{pointr} dataset is generated by noised back-projecting to better approximate the real scans, which has up to 55 shape categories and more train/test sets.
Besides, ViPC~\cite{vipc} proposes the first multimodal completion dataset ShapeNet-ViPC. It takes point clouds and the corresponding images as inputs, and is also a subset of the ShapeNet~\cite{shapenet} dataset with 13 categories.
However, due to the limited categories, single completion task, and the fully supervised training mode, the existing completion datasets cannot fully explore the diversity of shape categories, the robustness to noise, and the transfer ability to unseen categories of different point cloud completion methods.
{
\subsection{Multi-modal feature interaction}
The multi-modal feature interaction aims to comprehensively characterize the feature patterns of inputs, by utilizing complementary information from multiple perspectives and aspects mutually.
This feature interaction concept has been applied in various visual tasks, such as 3D semantic segmentation, object detection, depth completion and person Re-Identification.
In fact, in some LiDAR scenarios, cameras are widely present, and this additional image information hold significant importance for 3D applications.
PIF-Net\cite{PIFNET} introduces an encoder–decoder structure for multimodality semantic segmentation, which contains two independent branches with res-pooling and point attention blocks to extract deep and condensed multimodal features.
\cite{KP} introduces a novel key points extraction approach based on 2D image mapping. It first detects corner points in the image and maps them as auxiliary information to the 3D point cloud to achieve more accurate point cloud keypoint extraction.
To fully utilize the structural differences and the manual statistical characteristics between the multimodal very-high-resolution (VHR) aerial images and LiDAR data, MFNet\cite{MFNet} designed a novel deep multi-modal fusion network, which contains a multimodal fusion mechanism, pyramid dilation blocks and a multilevel feature fusion module.
For the multimodal feature fusion 3D detection task, MSF\cite{MSF} introduces an adaptive attention-based fusion module to fuse features by learning an adaptive fusion score, to tackle misalignment challenges between the 2D and 3D parsing results and boost the final detection performance. 
To our knowledge, our DuInNet is the first multi-modal feature interaction method in the point cloud completion task.
}
\section{Methodology}
\label{sec:DuInNet}
\begin{figure*}[t]
  \centering
  \subfigure[A schematic overview of the proposed DuInNet]{
  \begin{minipage}[t]{1\linewidth}
    \centering
    \includegraphics[width=1\linewidth]{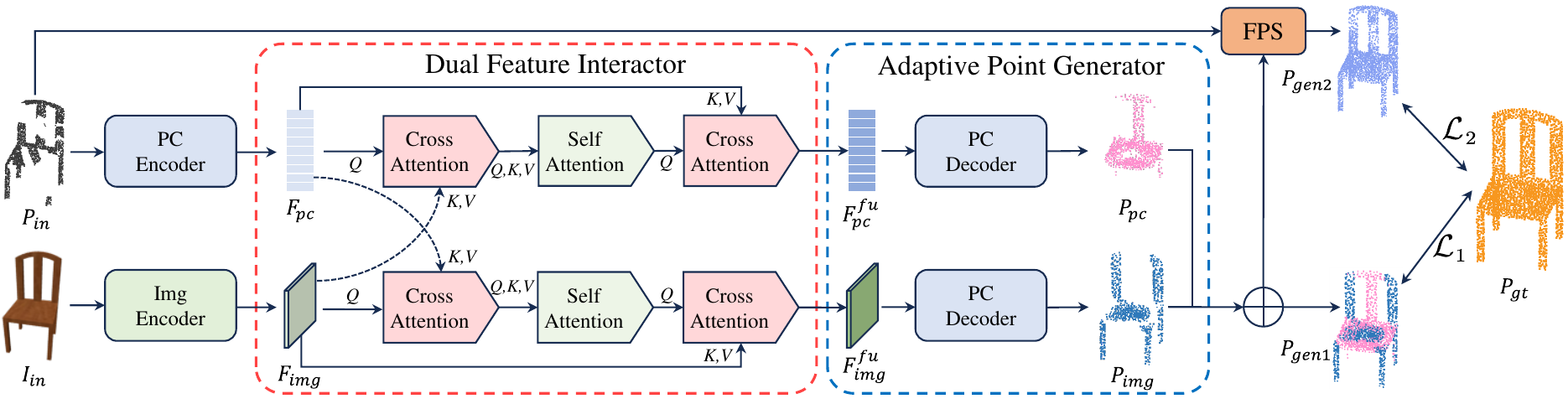}
    \end{minipage}%
    \label{fig:architecture1a}
  }
  \subfigure[The architecture of Dual Feature Interactor]{
  \begin{minipage}[t]{0.585\linewidth}
    \centering
    \includegraphics[width=1\linewidth]{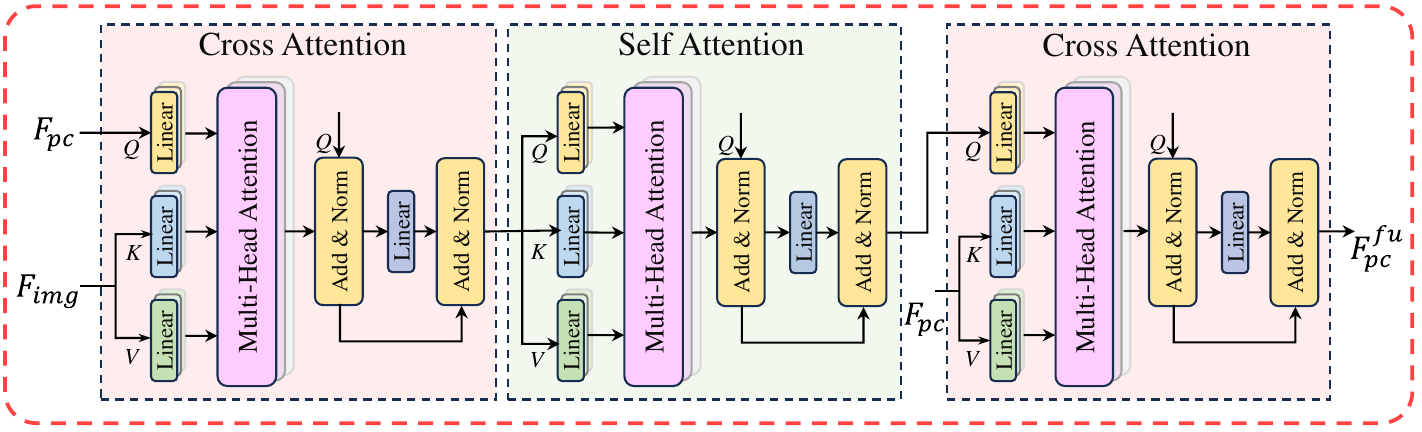}
    \end{minipage}%
    \label{fig:architecture1b}
  }
  \subfigure[The architecture of Point Cloud Decoder]{
  \begin{minipage}[t]{0.375\linewidth}
    \centering
    \includegraphics[width=1\linewidth]{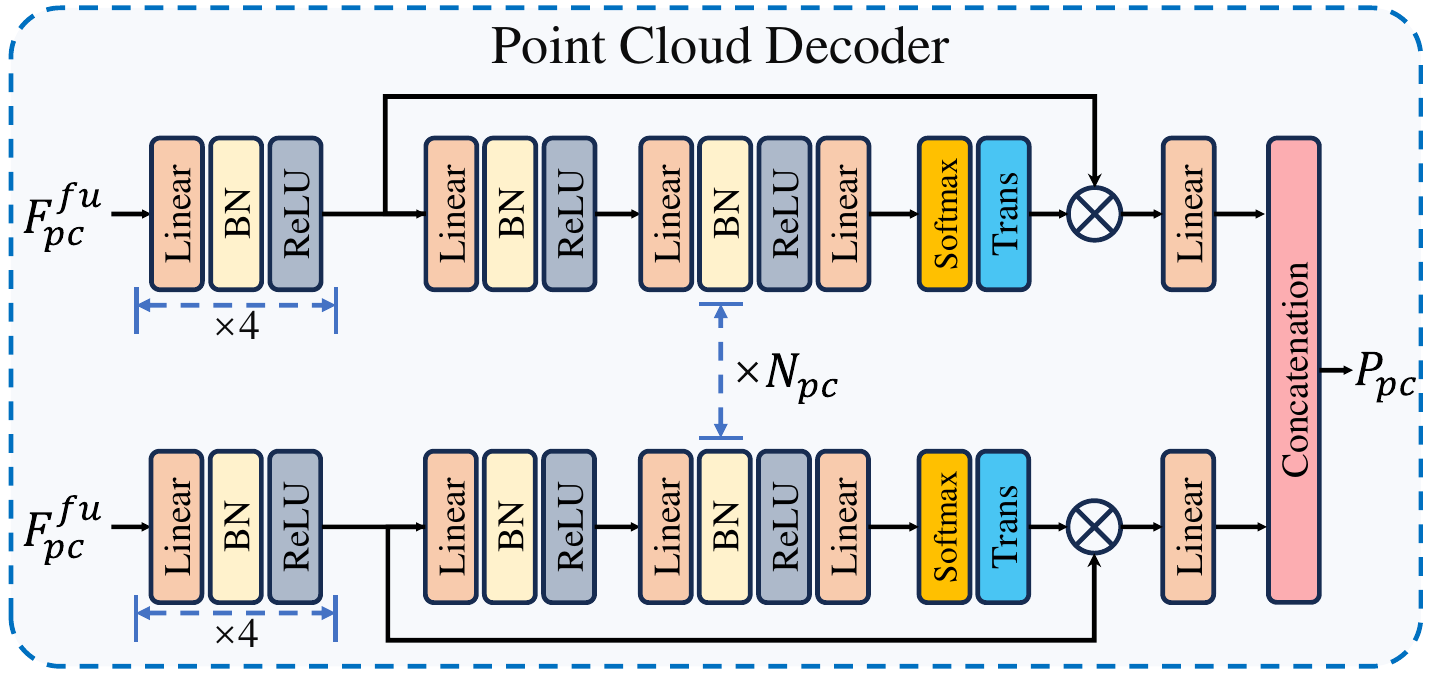}
    \end{minipage}%
    \label{fig:architecture1c}
  }
    \caption{The overall architecture of DuInNet. DuInNet adopts a dual-path autoencoder structure, takes partial point clouds and their corresponding 2D images as inputs, to generate complete point clouds. (b) and (c) take the point cloud path for example to illustrate the architectures. Here, FPS, BN and Trans represent farthest point sampling, batch normalization, and transpose, respectively.} 
   \label{fig:architecture}
\end{figure*}
\subsection{Overall Architecture}
The proposed DuInNet learns cross-modal feature representations and generates complete point clouds from two input modalities. As illustrated in Fig.~\ref{fig:architecture1a}, DuInNet adopts a dual-path autoencoder architecture, which consists of three core modules including point cloud and image encoders, dual feature interactor, and adaptive point generator. Given a partial point cloud and an image from the same object, the DuInNet interacts features between the point cloud and image, and adaptively reconstructs a complete point cloud from these two modalities.
{
It is worth noting that although multimodal completion datasets have paired view-aligned point clouds and images, it is unnecessary to use the paired data under the same viewpoint for point cloud completion.
Given a partial point cloud, multimodal completion methods randomly select a rendered image from all viewpoints with it as inputs to generate the complete shape, which is similar to real-world scenarios.}
Specifically, given a partial point cloud $P_{in} \in \mathbb{R}^{N \times 3}$ and its image $I_{in} \in \mathbb{R}^{H\times W \times 3}$, the role of DuInNet is to generate a new point cloud $P_{gen} \in \mathbb{R}^{N \times 3}$ to make it similar to the ground truth complete point cloud $P_{gt} \in \mathbb{R}^{N \times 3}$, with sufficient feature interaction between inputs $P_{in}$ and $I_{in}$. 

\subsection{Point Cloud and Image Encoders}

The point cloud and image encoders are used to learn pointwise local geometric features $F_{pc} \in \mathbb{R}^{N_{p} \times C}$ and pixelwise texture features $F_{img} \in \mathbb{R}^{N_{i} \times C}$, respectively. 

\subsubsection{Point cloud encoder}
To enlarge the receptive field, we embed two point transformer layers (PT)~\cite{pt} into the encoder of the PointNet++ (set abstraction layers with $k$ nearest neighbors, SAb)~\cite{pn++} as our pointwise feature learning backbone. 
{
Specifically, we upsample the partial point cloud $P_{in}$ to ensure that both the upsampled partial point cloud and ground truth point cloud $P_{gt}$ contain $N=2048$ points. 
We mark the feature dimensions of tensors in \textbf{boldface} and the detailed network architecture of the point cloud encoder is as follows:}

{
Upsampled partial point cloud $(2048\times\bm{3})$ $\rightarrow$ SAb$_1$ $(2048\times\bm{3}$ to $512\times\bm{128}, k=16)$ $\rightarrow$ PT$_1$$(512\times\bm{128}$ to $512\times\bm{128})$ $\rightarrow$ SAb$_2$$(512\times\bm{128}$ to $128\times\bm{256}, k=16)$ $\rightarrow$ PT$_2$$(128\times\bm{256}$ to $128\times\bm{256})$ $\rightarrow$ $F_{pc} \in \mathbb{R}^{128 \times \bm{256}}$.}

It is worth noting that, to maintain the dimension consistency between pointwise and pixelwise features for the convenience of subsequent cross attention operators, we remove the max-pooling layer at the end of PointNet++.

\subsubsection{Image encoder}
To balance the computational complexity with accuracy, we use ResNet18~\cite{resnet} as our encoder to extract features from the corresponding images. 
{
After that, the feature $F_{img} \in \mathbb{R}^{196 \times \bm{256}}$ has $N_i=14\times14=196$ pixels with $C=\bm{256}$ feature dimensions.}
\subsection{Dual Feature Interactor}
Although the point clouds and view-aligned images are rendered from the same object, they provide different properties of 3D shapes. For example, images contain rich texture information and clear shape boundaries, which are beneficial for the completion task. However, current methods~\cite{xmfnet,vipc,csdn} just utilize the images to assist point clouds, and cannot fully explore the benefits of images. Thus, in this paper, we propose a dual feature interactor to fully interact the features of these two modalities.

As shown in Fig.~\ref{fig:architecture1b}, given features $F_{pc}$ and $F_{img}$, the dual feature interactor adopts a dual-path architecture to iteratively facilitate interactions between point cloud and image features, in which one path emulates point distribution with fused point cloud features, and the other explores object textures with fused image features.
These two paths have the same structure but with different weighting parameters. To interact and fuse features between point clouds and images, each path is composed of a cross-attention layer, a self-attention layer, and a cross-attention layer.

Here, we take the point cloud path as an example to introduce the details of the dual feature interactor. For the first cross-attention layer (CA$_1$), the point cloud features $F_{pc}$ are treated as query tensors $Q_{pc}$, and the image features are treated as key tensors $K_{img}$ and value tensors $V_{img}$. Then, the point cloud features can aggregate features from various image locations by the cross-correlation between these two modalities. This process can be formulated as follows:
\begin{equation}
Q_{pc}=\Phi_q\left(F_{pc}\right), K_{img}=\Phi_k\left(F_{img}\right), V_{img}=\Phi_v\left(F_{img}\right),
\label{eq:first_ca1}
\end{equation}
\begin{equation}
F_{pc}^{mid}={\rm softmax}\left(\frac{Q_{pc}K_{img}^T}{\sqrt{C}}\right)V_{img},
\label{eq:first_ca2}
\end{equation}
\begin{equation}
F_{pc}^{norm}={\rm norm}\left(Q_{pc}+F_{pc}^{mid}\right),
\label{eq:first_ca3}
\end{equation}
\begin{equation}
F_{pc}^{ca}={\rm norm}\left(F_{pc}^{norm}+\Phi_{mid}\left(F_{pc}^{norm}\right)\right),
\label{eq:first_ca4}
\end{equation}
where $\Phi$s are served as feature mapping layers implemented by multi-layer perceptrons, $F_{pc}^{mid}$ and $F_{pc}^{norm}$ are intermediate features, and $C$ is the dimension of features. The generated features $F_{pc}^{ca} \in \mathbb{R}^{N_{p} \times C}$ at this layer can be viewed as point cloud features enriched by image features.

In the self-attention layer (SA), it works exactly like Eqs. (\ref{eq:first_ca1}) to Eqs. (\ref{eq:first_ca4}) to generate the updated feature $F_{pc}^{sa}$, except for that $Q$, $K$ and $V$ are all projected from the same feature $F_{pc}^{ca}$.
Similarly, in the last cross-attention layer (CA$_2$), $Q$ is projected from the updated feature $F_{pc}^{sa}$, $K$ and $V$ are projected from the incipient feature $F_{pc}$ to generate fused feature $F_{pc}^{fu}$. 
{
The detailed network architecture of the dual feature interactor is as follows:}

{
\textbf{Point cloud path.} Pointwise local geometric features $F_{pc} \in \mathbb{R}^{128 \times \bm{256}}$ $\rightarrow$ $F^{ca}_{pc}={\rm CA_1}(F_{pc},F_{img},F_{img})$ $\rightarrow$ $F^{sa}_{pc}={\rm SA}(F^{ca}_{pc},F^{ca}_{pc},F^{ca}_{pc})$ $\rightarrow$ $F^{fu}_{pc}={\rm CA_2}(F^{sa}_{pc},F_{pc},F_{pc})$ $\rightarrow$ point cloud fused feature $F^{fu}_{pc} \in \mathbb{R}^{128 \times \bm{256}}$.}

{
\textbf{Image path.} Pixelwise texture features $F_{img}\in\mathbb{R}^{196 \times \bm{256}}$ $\rightarrow$ $F^{ca}_{img}={\rm CA_1}(F_{img},F_{pc},F_{pc})$ $\rightarrow$ $F^{sa}_{img}={\rm SA}(F^{ca}_{img},F^{ca}_{img},F^{ca}_{img})$ $\rightarrow$ $F^{fu}_{img}={\rm CA_2}(F^{sa}_{img},F_{img},F_{img})$ $\rightarrow$ image fused feature $F^{fu}_{img} \in \mathbb{R}^{196 \times \bm{256}}$.}

\subsection{Adaptive Point Generator}
The adaptive point generator aims to transform the high-dimensional fused features into the coordinates of points in 3D space.
As shown in Fig.~\ref{fig:architecture1c}, it also adopts a dual-path structure to generate complete point clouds in blocks with different weights for two modalities according to the specific requirements of tasks.

For each block, we construct a transformation function $\mathcal M \left(\cdot\right)$ from fused features ($F_{pc}^{fu}$ or $F_{img}^{fu}$) to 3D coordinates ($P_{pc}^{i}$ or $P_{img}^{i} \in \mathbb{R}^{N_{block} \times 3}$) by using sequential linear, batch normalization and activation function layers (LBR). 
Each block contains $N_{block}=\frac{N}{N_{pc} + N_{img}}$ points.
{
We take the point cloud path for example, and the detailed network architecture of the transformation function $\mathcal M_i \left(\cdot\right)$ is as follows:}

{
\textbf{Step 1:} Point cloud fused feature $F^{fu}_{pc} \in \mathbb{R}^{128 \times \bm{256}}$ $\rightarrow$ LBR$_1$ $(128\times\bm{256}$ to $128\times\bm{512})$ $\rightarrow$ LBR$_2$$(128\times\bm{512}$ to $128\times\bm{512})$ $\rightarrow$ LBR$_3$$(128\times\bm{512}$ to $128\times\bm{512})$ $\rightarrow$ LBR$_4$$(128\times\bm{512}$ to $128\times\bm{64})$ $\rightarrow$ point cloud mapping feature $F^{map}_{pc} \in \mathbb{R}^{128 \times \bm{64}}$.}

{
\textbf{Step 2:} Point cloud mapping feature $F^{map}_{pc}\in\mathbb{R}^{128\times\bm{64}}$$\rightarrow$ LBR$_5$ $(128\times\bm{64}$ to $128\times\bm{64})$ $\rightarrow$ LBR$_6$ $(128\times\bm{64}$ to $128\times\bm{128})$ $\rightarrow$ Linear$_1$$(128\times\bm{128}$ to $128\times\bm{128})$ $\rightarrow$ Transpose$(128\times\bm{128}$ to $\bm{128}\times128)$ $\rightarrow$ point cloud shape weight $F^{w}_{pc} \in \mathbb{R}^{\bm{128} \times 128}$.}

{
\textbf{Step 3:} $P_{pc}^{i}\in \mathbb{R}^{128 \times \bm{3}}$ = $\mathcal M_i (F_{pc}^{fu})$ = Linear$_2$ $(F^{w}_{pc} \times F^{map}_{pc})$. Here, Linear$_2$ $(\bm{128}\times\bm{64}$ to $128\times\bm{3})$.}

{
Similarly, in the image path, we utilize the function $\mathcal M_j \left(\cdot\right)$ to transform fused features $F_{img}^{fu}\in \mathbb{R}^{196 \times \bm{256}}$ into 3D coordinates $P_{img}^{j} \in \mathbb{R}^{128 \times \bm{3}}$.}

Thus, following the above paradigm, the adaptive point generator generates ${N_{pc}}$ block point clouds by the point cloud path, and ${N_{img}}$ block point clouds by the image path. Finally, we concatenate the point cloud of each block to form the restored complete point cloud $P_{gen1}\in \mathbb{R}^{N \times 3}$, which has the same number of points as $P_{gt}$. The process can be formulated as follows:
\begin{equation}
\begin{cases}
P_{pc}^{i} = \mathcal M_i \big(F_{pc}^{fu}\big)  &\quad i=1,\cdot\cdot\cdot,N_{pc} \\
P_{img}^{j} = \mathcal M_j \big(F_{img}^{fu}\big)  &\quad j=1,\cdot\cdot\cdot,N_{img} \\
\end{cases}
\label{eq:apg1}
\end{equation}
\begin{equation}
P_{gen1}={\rm Cat}\{P_{pc}^{1},\cdot\cdot\cdot,P_{pc}^{N_{pc}},P_{img}^{1},\cdot\cdot\cdot,P_{img}^{N_{img}}\},
\label{eq:apg2}
\end{equation}

In addition, we concatenate $P_{gen1}$ with the partial input point cloud, and generate another complete point cloud $P_{gen2}\in \mathbb{R}^{N \times 3}$ with the Farthest Point Sampling (FPS) algorithm. This process is formulated as follows: 
\begin{equation}
P_{gen2}={\rm FPS}\left({\rm Cat}\{P_{in},P_{gen1}\}\right).
\label{eq:apg3}
\end{equation}

\begin{figure}[t]
  \centering
  \subfigure[CAD model]{
  \begin{minipage}[t]{0.23\linewidth}
    \centering
    \includegraphics[width=1\linewidth]{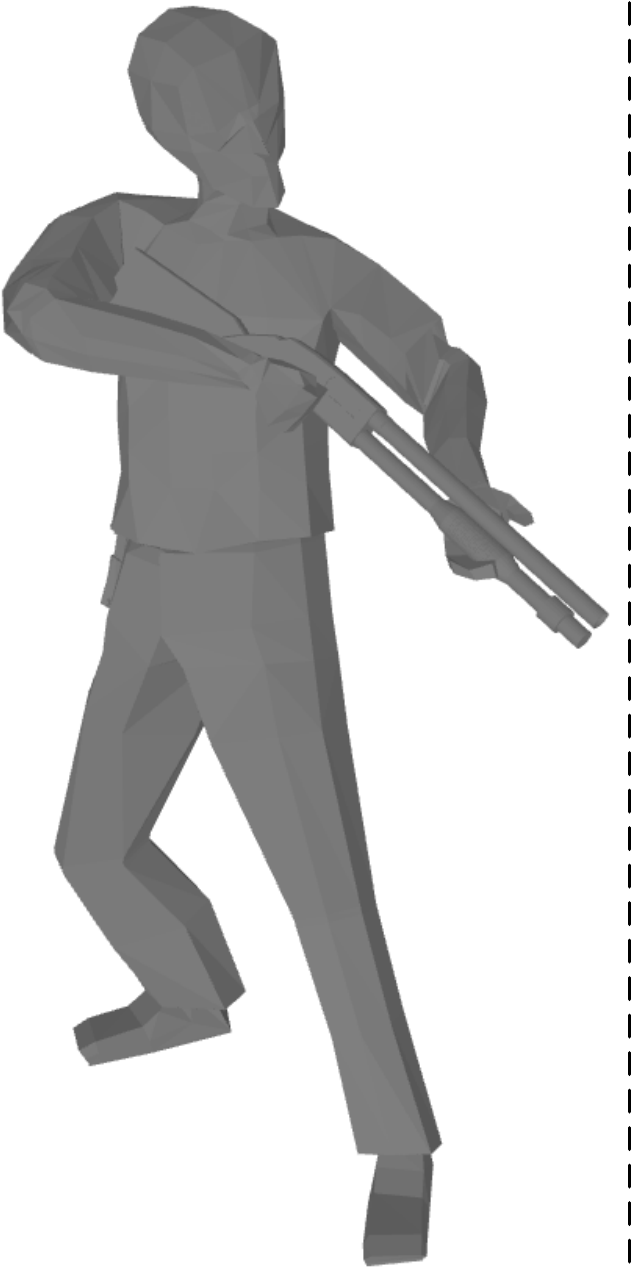}
    \end{minipage}
    \label{fig:pds1}
  }
  \subfigure[Random]{
  \begin{minipage}[t]{0.195\linewidth}
    \centering
    \includegraphics[width=1\linewidth]{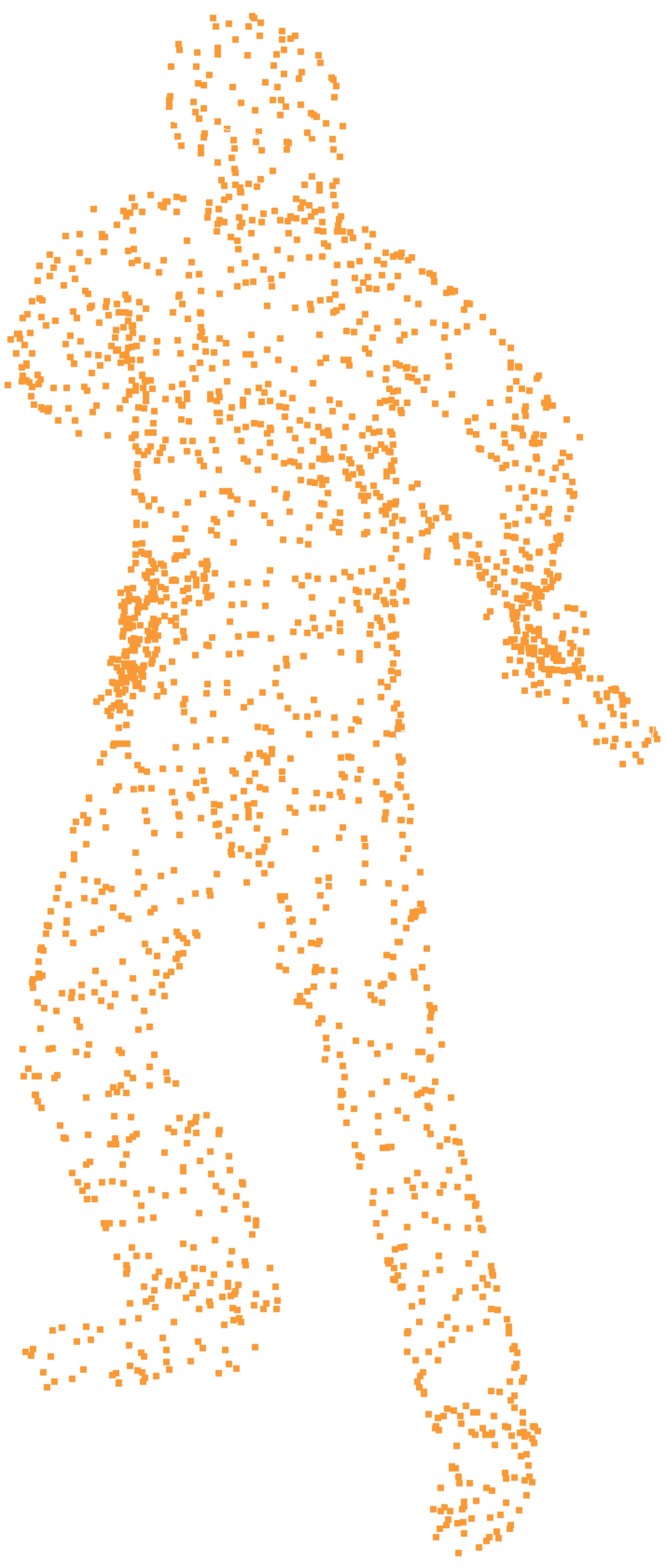}
    \end{minipage}
    \label{fig:pds2}
  }
  \subfigure[Uniform]{
  \begin{minipage}[t]{0.195\linewidth}
    \centering
    \includegraphics[width=1\linewidth]{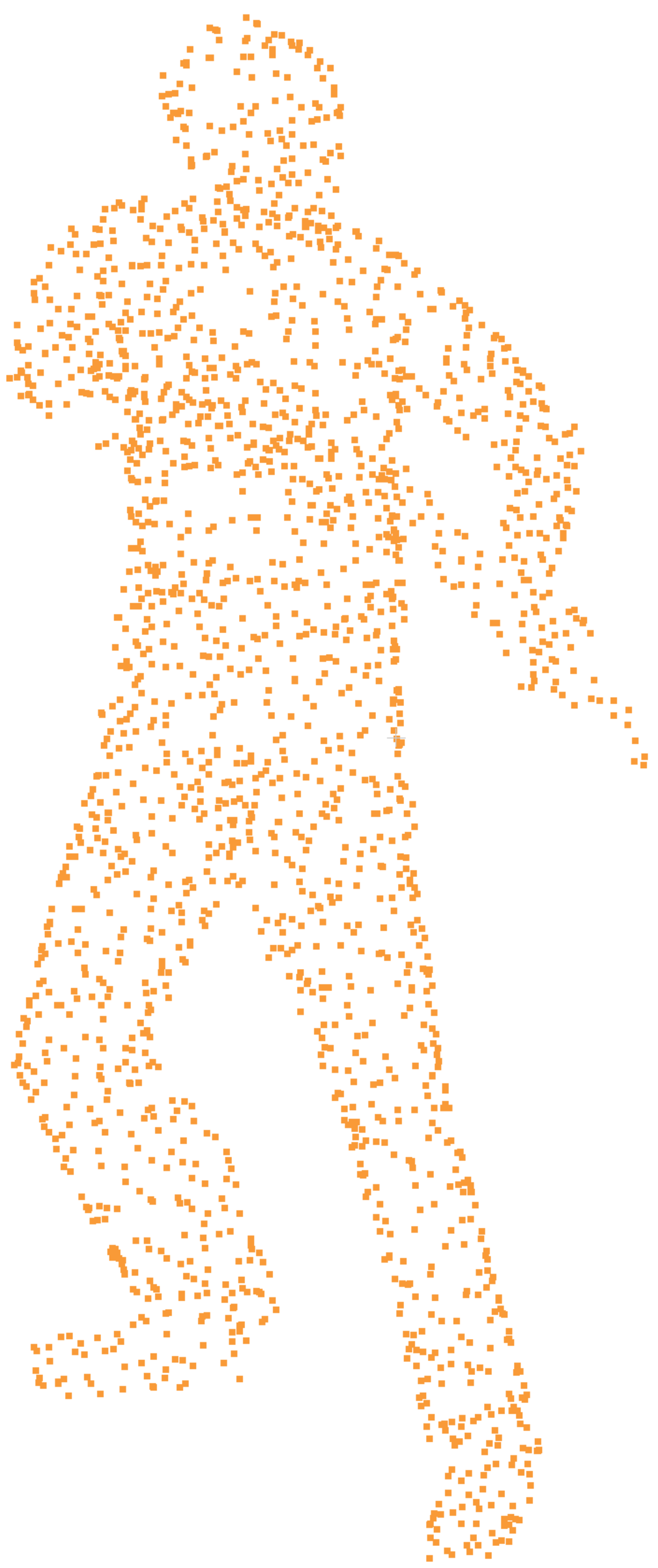}
    \end{minipage}
    \label{fig:pds3}
  }
  \subfigure[PDS]{
  \begin{minipage}[t]{0.195\linewidth}
    \centering
    \includegraphics[width=1\linewidth]{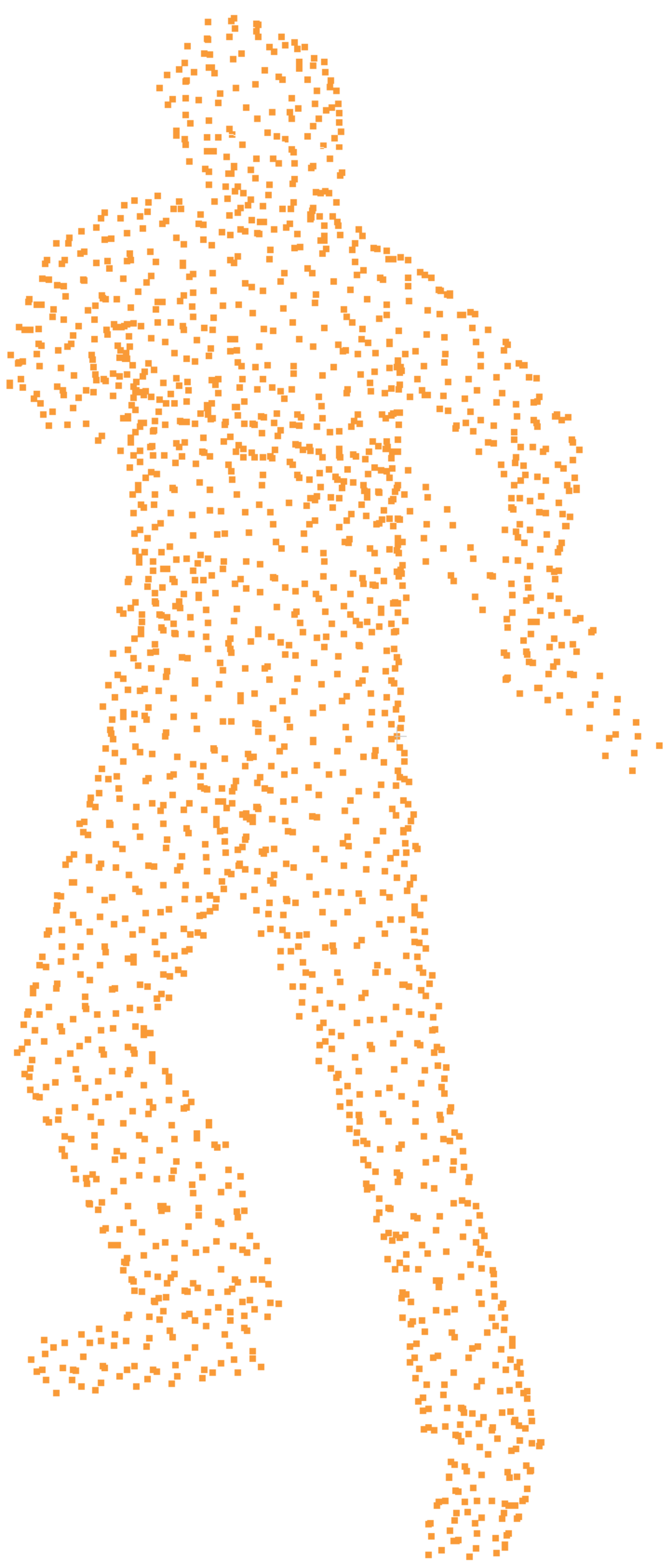}
    \end{minipage}
    \label{fig:pds4}
  }
   \caption{{Qualitative comparisons of different point cloud sampling strategies. (a) a CAD model of the person category. (b), (c) and (d) are corresponding point clouds rendered by randomly sampling, uniformly sampling, and poisson disk sampling strategies, respectively.}}
   \label{fig:pds}
\end{figure}

\subsection{Loss Function}

In our method, during the training phase, we consider the losses from two generated complete point clouds, and utilize the $\ell_1$-norm chamfer distance ($d_{CD}^{\ell_1}$) as the loss function: 
\begin{equation}
\begin{aligned}
\mathcal{L}&= \mathcal{L}_{CD1}+\mathcal{L}_{CD2} \\
&=d_{CD}^{\ell_1}(P_{gen1},P_{gt})+d_{CD}^{\ell_1}(P_{gen2},P_{gt}),
\label{eq:cd2}
\end{aligned}
\end{equation}
{
\begin{equation}
\begin{aligned}
d_{CD}^{\ell_1}(\mathcal{P}, \mathcal{Q})= &\frac{1}{2|\mathcal{P}|} \sum_{p \in \mathcal{P}} \min _{q \in \mathcal{Q}}\sqrt{\|p-q\|}  \\
&+\frac{1}{2|\mathcal{Q}|} \sum_{q \in \mathcal{Q}} \min _{p \in \mathcal{P}}\sqrt{\|q-p\|},
\end{aligned}
\label{eq:cdl1}
\end{equation}}
where $\mathcal{P}$ and $\mathcal{Q}$ are two point clouds, and $p$, $q$ are points in $\mathcal{P}$ and $\mathcal{Q}$, respectively.

Considering that noisy partial point clouds have adverse effects on network training, we only use  $\mathcal{L}_{CD1}$ as the loss function to train the network in the denoising point cloud completion task. For other tasks, we use $\mathcal{L}_{CD}$ as the loss function.
\section{ModelNet-MPC Dataset}
\label{sec:ModelNet-MPC}

\subsection{Motivation}
To provide a benchmark for extensive investigations of point cloud completion methods in terms of noise robustness, category generalization and transfer ability, we introduce ModelNet-MPC, a high-quality multimodal and multi-view point cloud completion dataset.
We compare the ModelNet-MPC dataset with several existing point cloud completion benchmarks in Table~\ref{tab:dataset}, including PCN~\cite{pcn}, Completion3D~\cite{topnet}, MVP~\cite{vrcnet}, ShapeNet-55~\cite{pointr} and ShapeNet-ViPC~\cite{vipc}. It is clear that the ModelNet-MPC dataset has several advantages over all existing datasets. 
\begin{table*}[t]
\caption{\textbf{Comparative results achieved on existing datasets.} ModelNet-MPC has several appealing properties, such as: 1) multiple modalities; 2) rich categories; 3) more test data; 4) high-quality point clouds; 5) various expanded tasks. (C3D: Completion3D; MVP: Multi-View Partial; PC: Point clouds; Img: Images; Cat.: Categories; Sam.: Sampling; PDS: Poisson Disk Sampling; Num.: Numbers; Distri.: Distribution; Reso.: Resolution.) The optimal and suboptimal properties are highlighted in \textbf{boldface} and \uline{underlined}, respectively.}
\centering
\renewcommand\arraystretch{1.2}
\resizebox{\textwidth}{!}{
\begin{tabular}{c|c|c|cc|cc|c|cc|c|c|c}
\hline
 &
   &
   &
  \multicolumn{2}{c|}{Training Set} &
  \multicolumn{2}{c|}{Testing Set} &
   &
  \multicolumn{2}{c|}{Virtual View} &
   &
   &
   \\
\multirow{-2}{*}{} &
  \multirow{-2}{*}{Modalities} &
  \multirow{-2}{*}{\#Cat.} &
  \#CAD &
  \#Pair &
  \#CAD &
  \#Pair &
  \multirow{-2}{*}{PC Sam.} &
  Num. &
  Distri. &
  \multirow{-2}{*}{Img Reso.} &
  \multirow{-2}{*}{\begin{tabular}[c]{@{}c@{}}Denoising\\ Task\end{tabular}} &
  \multirow{-2}{*}{\begin{tabular}[c]{@{}c@{}}One-Shot\\ Task\end{tabular}} \\ \hline\hline
PCN~\cite{pcn} &
  PC &
  8 &
  28974 &
  231792 &
  1200 &
  1200 &
  Uniform &
  8 &
  Random &
  \ding{55} &
  \ding{55} &
  \ding{55} \\
C3D~\cite{topnet} &
  PC &
  8 &
  28974 &
  28974 &
  1184 &
  1184 &
  Uniform &
  1 &
  Random &
  \ding{55} &
  \ding{55} &
  \ding{55} \\
MVP~\cite{vrcnet} &
  PC &
  16 &
  2400 &
  62400 &
  1600 &
  41600 &
  \textbf{PDS} &
  \uline{26} &
  \textbf{Uniform} &
  \ding{55} &
  \ding{55} &
  \ding{55} \\
ShapeNet-55~\cite{pointr} &
  PC &
  \textbf{55} &
  \textbf{41952} &
  \uline{335616} &
  \uline{10518} &
  84144 &
  Random &
  8 &
  \textbf{Uniform} &
  \ding{55} &
 \ding{52} &
  \ding{52} \\ \hline
ShapeNet-ViPC~\cite{vipc} &
  \textbf{PC+Img} &
  13 &
  \uline{31650} &
  \textbf{607575} &
  \textbf{31503} &
\uline{152025} &
  Random &
  24 &
  \textbf{Uniform} &
  137 $\times$ 137  &
  \ding{55} &
  \ding{55} \\ 
\rowcolor[HTML]{EFEFEF} 
\textbf{ModelNet-MPC} (Ours) &
  \textbf{PC+Img} &
  \uline{40} &
  6156 &
  196992 &
  \cellcolor[HTML]{EFEFEF}6155 &
  \textbf{196960} &
  \textbf{PDS} &
  \textbf{32} &
  \textbf{Uniform} &
  \textbf{224} ${\bf{\times}}$ \textbf{224} &
  \ding{52} &
  \ding{52} \\ \hline
\end{tabular}
}
\label{tab:dataset}
\end{table*}
\begin{figure*}[t]
  \centering
  \subfigure[32 uniform poses]{
  \begin{minipage}[t]{0.145\linewidth}
    \centering
    \includegraphics[width=1\linewidth]{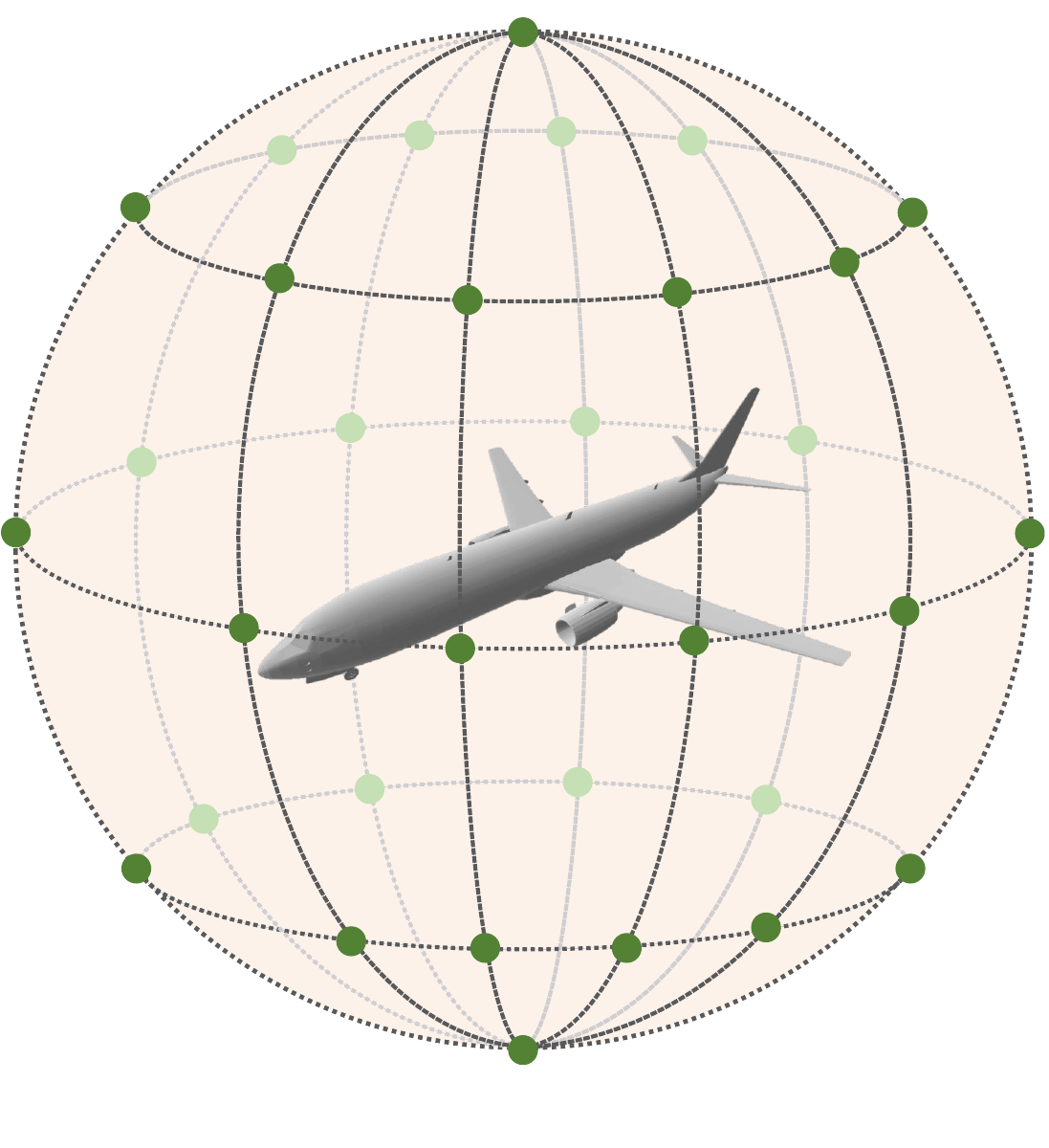}
    \end{minipage}
    \label{fig:dataset1a}
  }
  \subfigure[The 32 rendered images for this airplane CAD]{
  \begin{minipage}[t]{0.337\linewidth}
    \centering
    \includegraphics[width=1\linewidth]{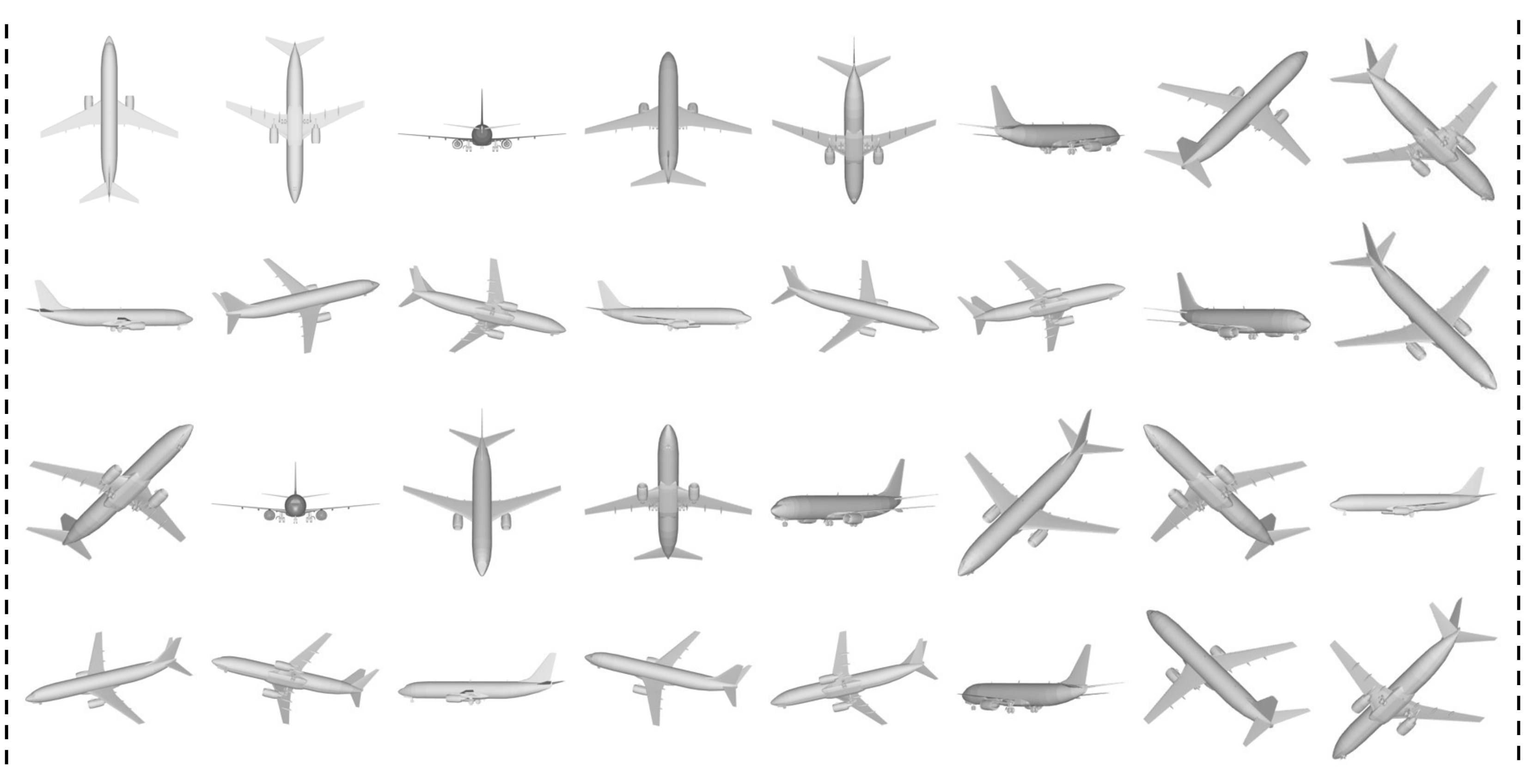}
    \end{minipage}
    \label{fig:dataset1b}
  }
  \subfigure[The 32 rendered partial point clouds for this airplane CAD]{
  \begin{minipage}[t]{0.465\linewidth}
    \centering
    \includegraphics[width=1\linewidth]{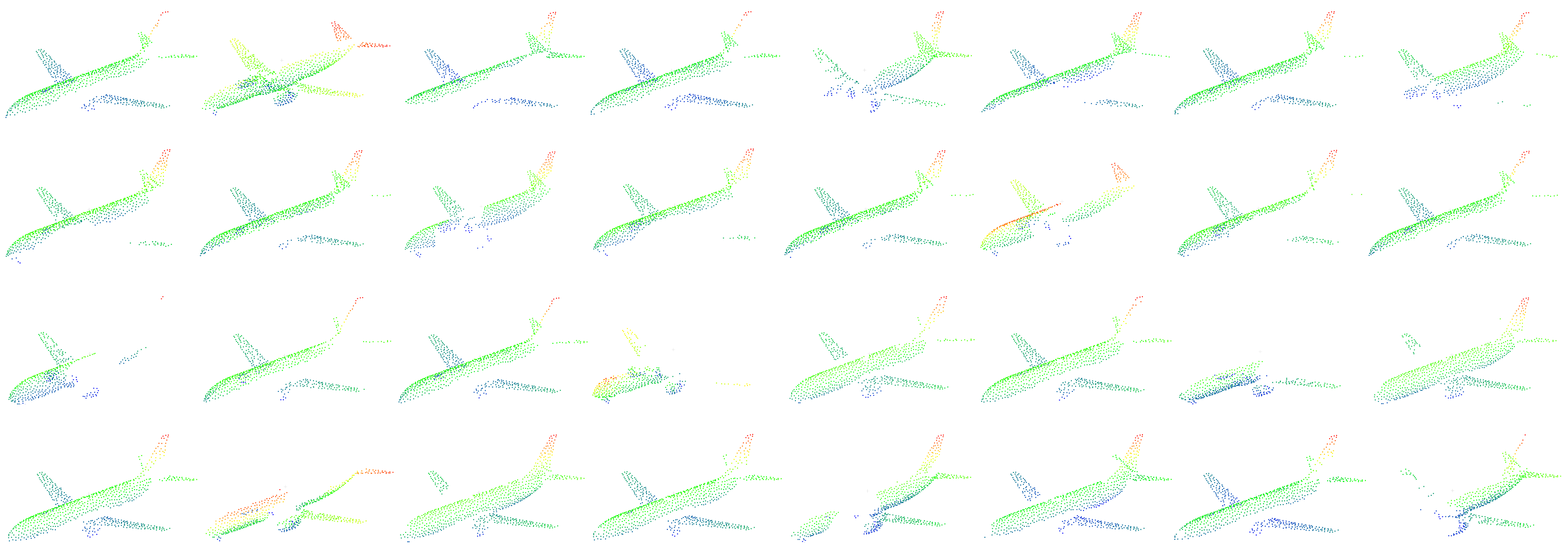}
    \end{minipage}
    \label{fig:dataset1c}
  }
   \caption{Our \textbf{ModelNet-}based \textbf{M}ultimodal \textbf{P}oint cloud \textbf{C}ompletion dataset (\textbf{ModelNet-MPC}). \textbf{(a)} shows 32 uniformly distributed viewpoints on a unit sphere. \textbf{(b)} and \textbf{(c)} represent the 32 {view-aligned} rendered partial point clouds and images for the airplane CAD from left uniformly distributed viewpoints, respectively.} 
   \label{fig:dataset}
\end{figure*}

\subsection{CAD Model Rendering}
We build our ModelNet-MPC dataset based on the ModelNet40 dataset~\cite{modelnet}, which contains 12,311 objects from 40 categories. 
The ModelNet-MPC dataset not only contains common object categories such as airplane, chair, and sofa, but also contains some rare categories, such as piano, tent, and xbox.
For each CAD model, we generate three sets of data:
\subsubsection{Complete point clouds}
{
One way to render point clouds from object CAD models (Fig.~\ref{fig:pds1}) is to simply sample their surfaces randomly (Fig.~\ref{fig:pds2}) or uniformly (Fig.~\ref{fig:pds3}). 
Although fast and easy to implement, both methods result in unsatisfying messy point clouds.
To solve this problem, the poisson disk sampling algorithm (PDS)~\cite{pds} is proposed (Fig.~\ref{fig:pds4}). 
Although the points generated by PDS are tightly packed together, the distance between them does not exceed a specified minimum distance.
As shown in Fig.~\ref{fig:pds}, PDS can generate smoother complete point clouds than random and uniform sampling, making them a better representation of different object CAD models.
Therefore, our ModelNet-MPC benchmark employs the poisson disk sampling algorithm~\cite{pds} to sample 2048 points from each CAD surface as complete point clouds, to evaluate capabilities of restoring high-quality geometric details between different point cloud completion methods.}

\subsubsection{Images rendered from different viewpoints} As shown in Fig.~\ref{fig:dataset1a}, we uniformly select 32 fixed viewpoints (for camera poses) on a unit sphere for each CAD model.
For each fixed viewpoint, we set up a virtual camera facing towards the center of the object and render an image with a resolution of $224\times224$ (as shown in Fig.~\ref{fig:dataset1b}).
\subsubsection{Two types of partial point clouds}
{
Most point clouds captured by existing 3D sensors are usually incomplete and noisy, due to the limits of objects self-occlusion and sensors viewing angles.
The hidden point removal (HPR) algorithm~\cite{hpr} proposes a simple and fast operator to determine the visible points in a point cloud, as viewed from a given viewpoint.}
Therefore, building upon the previously selected viewpoints, we adopt the HPR algorithm ~\cite{hpr} to generate partial point clouds (as shown in Fig.~\ref{fig:dataset1c}) for the ModelNet-MPC, with the ratios of point numbers varying from 10\% to 40\% of the complete point cloud.
Besides, we add Gaussian noise to the generated partial point clouds to create noisy point clouds, which are used to evaluate the denoising completion task.

\subsection{Appealing Properties}
ModelNet-MPC has more appealing properties than other datasets: 
\subsubsection{Multiple modalities} ModelNet-MPC is the second multimodal point cloud completion dataset, which contains nearly 400,000 pairs of high-quality point clouds and {rendered} images, and will boost further developments in this field.
\subsubsection{Rich categories} ModelNet-MPC contains 40 categories, and is more than three times of ShapeNet-ViPC~\cite{vipc}. Such rich object categories are beneficial for the test of the category generalization ability of different completion methods.
\subsubsection{More diverse test data} ModelNet-MPC creates a challenging benchmark and increases the proportion of test samples from 20\% to 50\%. 
More importantly, unlike ShapeNet-ViPC which randomly divides the training and test data according to viewpoints, ModelNet-MPC strictly divides them according to object categories, ensuring that networks have never been trained on instances with their corresponding complete point clouds in the test dataset.
\subsubsection{High-quality point clouds} ModelNet-MPC samples complete point clouds with 2048 points from each CAD surface by the poisson disk sampling method~\cite{pds}, which are smoother and retain more detailed structures than the point clouds obtained using the uniform sampling method.
\subsubsection{Various expanded tasks} Compared with ShapeNet-ViPC, ModelNet-MPC incorporates two additional tasks including completion from the denoising and zero-shot learning, providing more comprehensive evaluations for different methods.

\subsection{Expanded Tasks}
In addition to the fully supervised point cloud completion task, ModelNet-MPC also introduces two novel expanded {completion} tasks:
\subsubsection{The denoising {completion} task} Noisy point clouds are common in real world for data captured by 3D sensors. To simulate real-world scenarios, we add some Gaussian noises to the clean partial point clouds, and generate a new set of more challenging partial point clouds. By introducing the denoising task to ModelNet-MPC, we expect that our dataset can be beneficial for more robust and advanced point cloud completion models.   
\subsubsection{The zero-shot learning {completion} task} Considering the diversity of object categories in the real world, completion methods should also possess good transfer performance for unknown object categories, which is particularly crucial for practical applications.
Therefore, we introduce the zero-shot learning {completion} task to ModelNet-MPC. Specifically, we partition the ModelNet-MPC dataset into two subsets according to object categories. To evaluate the zero-shot transfer ability of different methods, we train models with 30 categories and test models with 40 categories (including 10 unseen categories{: bowl, cup, curtain, keyboard, radio, sink, stairs, stool, tent and wardrobe}).
\section{Experiments}
\label{sec:experiments}
\begin{table*}[t]
\caption{Quantitative comparisons to the state-of-the-art methods on eight categories of the ShapeNet-ViPC dataset in terms of Mean Chamfer Distance per point (CD, $\times 10^{-3}$) and Mean F-Score @ 0.001 (FS). Lower CD and higher FS are better. The best results are highlighted in \textbf{boldface}.}
\centering
\renewcommand\arraystretch{1.15}
\resizebox{\textwidth}{!}{
\begin{tabular}{c|c|r||cc||cc|cc|cc|cc|cc|cc|cc|cc}
\hline
\multicolumn{1}{l|}{} &
  \multirow{2}{*}{Method} &
  \multirow{2}{*}{Reference\;\;} &
  \multicolumn{2}{c||}{Mean} &
  \multicolumn{2}{c|}{Airplane} &
  \multicolumn{2}{c|}{Cabinet} &
  \multicolumn{2}{c|}{Car} &
  \multicolumn{2}{c|}{Chair} &
  \multicolumn{2}{c|}{Lamp} &
  \multicolumn{2}{c|}{Sofa} &
  \multicolumn{2}{c|}{Table} &
  \multicolumn{2}{c}{Watercraft} \\
 &
   &
   &
  CD&
  FS&
  CD&
  FS&
  CD&
  FS&
  CD&
  \multicolumn{1}{c|}{FS} &
  CD&
  FS&
  CD&
  FS&
  CD&
  FS&
  CD&
  FS&
  CD&
  FS \\ \hline  \hline 
\multirow{12}{*}{\rotatebox{90}{Single-modal}} &
  AtlasNet~\cite{atlasnet} & CVPR 2018
   &
  6.062&
  0.410&
  5.032&
  0.509&
  6.414&
  0.304&
  4.868&
  0.379&
  8.161&
  0.326&
  7.182&
  0.426&
  6.023&
  0.318&
  6.561&
  0.469&
  4.261&
  0.551\\
 &
  FoldingNet~\cite{foldingnet} &CVPR 2018
   &
  6.271&
  0.331&
  5.242&
  0.432&
  6.958&
  0.237&
  5.307&
  0.300&
  8.823&
  0.204&
  6.504&
  0.360&
  6.368&
  0.249&
  7.080&
  0.351&
  3.882&
  0.518\\
 &
  PCN~\cite{pcn} &3DV 2018
   &
  5.619&
  0.407&
  4.246&
  0.578&
  6.409&
  0.270&
  4.840&
  0.331&
  7.441&
  0.323&
  6.331&
  0.456&
  5.668&
  0.293&
  6.508&
  0.431&
  3.510&
  0.577\\
 &
  TopNet~\cite{topnet}&CVPR 2019
   &
  4.976&
  0.467&
  3.710&
  0.593&
  5.629&
  0.358&
  4.530&
  0.405&
  6.391&
  0.388&
  5.547&
  0.491&
  5.281&
  0.361&
  5.381&
  0.528&
  3.350&
  0.615\\
 &
  VRC-Net~\cite{vrcnet}&CVPR 2021
   &
  4.598&
  0.764&
  2.813&
  0.902&
  6.108&
  0.621&
  4.932&
  0.753&
  5.342&
  0.722&
  4.103&
  0.823&
  6.614&
  0.654&
  3.953&
  0.810&
  2.925&
  0.832\\
 &
  PF-Net~\cite{pfnet}&CVPR 2020
   &
  3.873&
  0.551&
  2.515&
  0.718&
  4.453&
  0.399&
  3.602&
  0.453&
  4.478&
  0.489&
  5.185&
  0.559&
  4.113&
  0.409&
  3.838&
  0.614&
  2.871&
  0.656\\
 &
  MSN~\cite{msn}&AAAI 2020
   &
  3.793&
  0.578&
  2.038&
  0.798&
  5.060&
  0.378&
  4.322&
  0.380&
  4.135&
  0.562&
  4.247&
  0.652&
  4.183&
  0.410&
  3.976&
  0.615&
  2.379&
  0.708\\
 &
  GRNet~\cite{grnet}&ECCV 2020
   &
  3.171&
  0.601&
  1.916&
  0.767&
  4.468&
  0.426&
  3.915&
  0.446&
  3.402&
  0.575&
  3.034&
  0.694&
  3.872&
  0.450&
  3.071&
  0.639&
  2.160&
  0.704\\
 &
  PoinTr~\cite{pointr}&ICCV 2021
   &
  2.851&
  0.683&
  1.686&
  0.842&
  4.001&
  0.516&
  3.203&
  0.545&
  3.111&
  0.662&
  2.928&
  0.742&
  3.507&
  0.547&
  2.845&
  0.723&
  1.737&
  0.780\\
 &
  SnowflakeNet~\cite{snowflakenet} &TPAMI 2022
   &
  1.450 &
  0.799 &
  0.644 &
  0.952 &
  2.093 &
  0.672 &
  2.041 &
  0.652 &
  1.480 &
  0.803 &
  0.810 &
  0.909 &
  1.916 &
  0.700 &
  1.695 &
  0.801 &
  0.920 &
  0.900 \\
 &
  Seedformer~\cite{seedformer}&ECCV 2022
   &
  2.902&
  0.688&
  1.716&
  0.835&
  4.049&
  0.551&
  3.392&
  0.544&
  3.151&
  0.668&
  3.226&
  0.777&
  3.603&
  0.555&
  2.803&
  0.716&
  1.679&
  0.786 \\ \hline
  
\multirow{4}{*}{\rotatebox{90}{Multi-modal}} &
  ViPC~\cite{vipc} &CVPR 2021
   &
  3.308&
  0.591&
  1.760&
  0.803&
  4.558&
  0.451&
  3.183&
  0.512&
  2.476&
  0.529&
  2.867&
  0.706&
  4.481&
  0.434&
  4.990&
  0.594&
  2.197&
  0.730 \\
 &
  CSDN~\cite{csdn}&TVCG 2023
   &
  2.570&
  0.695&
  1.251&
  0.862&
  3.670&
  0.548&
  2.977&
  0.560&
  2.835&
  0.669&
  2.554&
  0.761&
  3.240&
  0.557&
  2.575&
  0.729&
  1.742&
  0.782\\
 &
  XMFNet~\cite{xmfnet}&NeurIPS 2022
   &
  1.443&
  0.796&
  0.572&
  0.961&
  1.980&
  0.662&
  1.754&
  0.691&
  1.403&
  0.809&
  1.810&
  0.792&
  1.702&
  0.723&
  1.386&
  0.830&
  0.945&
  0.901 \\
 & 
  \textbf{DuInNet (Ours)} &/\quad\quad\;\,
   &
  \textbf{1.147}&
  \textbf{0.851}&
  \textbf{0.545}&
  \textbf{0.971}&
  \textbf{1.667}&
  \textbf{0.745}&
  \textbf{1.533}&
  \textbf{0.753}&
  \textbf{1.153}&
  \textbf{0.858}&
  \textbf{0.770}&
  \textbf{0.919}&
  \textbf{1.526}&
  \textbf{0.770}&
  \textbf{1.217}&
  \textbf{0.857}&
  \textbf{0.765}&
  \textbf{0.935}\\ \hline
\end{tabular}}
\label{tab:shapenetvipc}
\end{table*}
{
In this section, we first introduce the evaluation metrics and implementation details. Then, we compare our DuInNet to several state-of-the-art point cloud completion methods in both ShapeNet-ViPC and ModelNet-MPC datasets. Finally, we present ablation studies to investigate our network.}
{
\subsection{Evaluation Metrics}
We evaluate the point cloud completion performance of different methods with $\ell_2$-norm chamfer distance ($d_{CD}^{\ell_2}$) and F-Score (FS) metrics~\cite{fs}. They are defined as follows:
\begin{equation}
\begin{aligned}
d_{CD}^{\ell_2}(\mathcal{P}, \mathcal{Q}) = &\frac{1}{|\mathcal{P}|} \sum_{p \in \mathcal{P}} \min _{q \in \mathcal{Q}}\|p-q\| \\
&+ \frac{1}{|\mathcal{Q}|} \sum_{q \in \mathcal{Q}} \min _{p \in \mathcal{P}}\|q-p\|,
\end{aligned}
\label{eq:cdl2}
\end{equation}
where $\mathcal{P}$ and $\mathcal{Q}$ are two point clouds, and $p$, $q$ are points in $\mathcal{P}$ and $\mathcal{Q}$, respectively.}

{
As for the F-Score, we define the precision and recall of the point cloud completion task with the threshold $d$ as follows:
\begin{equation}
P(d) =  \frac{1}{|\mathcal{P}|}\sum_{p\in \mathcal{P}}  [\min_{q\in \mathcal{Q}}  ||p-q|| < d],
\label{eq:fs1}
\end{equation}
\begin{equation}
R(d) = \frac{1}{|\mathcal{Q}|}\sum_{q\in \mathcal{Q}} \lbrack \min_{p\in \mathcal{P}} ||q-p|| < d].
\label{eq:fs2}
\end{equation}}
{
Then, the F-Score@$d$ are defined as:
\begin{equation}
\text{F-Score}(d) = \frac{2P(d)R(d)}{P(d)+R(d)}, 
\label{eq:fs3}
\end{equation}
where $P(d)$ and $R(d)$ denote the precision and recall. In our experiments, we set the threshold $d$ to 0.001.}

\subsection{Implementation Details}
We evaluate our DuInNet on the ShapeNet-ViPC~\cite{vipc} and ModelNet-MPC datasets. We upsample the partial point cloud $P_{in}$ to make sure that the upsampled partial point cloud and ground truth point cloud $P_{gt}$ both contain $N=2048$ points. We also upsample input images $I_{in}$ of ShapeNet-ViPC to make them have the same size of $224\times224$ pixels as ModelNet-MPC.
In the point cloud encoder, the two set abstraction layers use $k=16$ nearest neighbors, and downsample the input point clouds $P_{in}$ by a factor of 16. Consequently, $F_{pc}$ has $N_p=128$ points and $C=256$ feature dimensions.
In the image encoder, the feature $F_{img}$ extracted by ResNet18~\cite{resnet} have $N_i=14\times14=196$ pixels with $C=256$ feature dimensions.
In the dual feature interactor, each multi-head attention has 4 heads.
In the adaptive point generator, each complete point cloud is composed of $N_{pc}+N_{img}=16$ blocks, while each block produces $128$ points.

\begin{figure*}[t]
   \centering
   \includegraphics[width=1\linewidth]{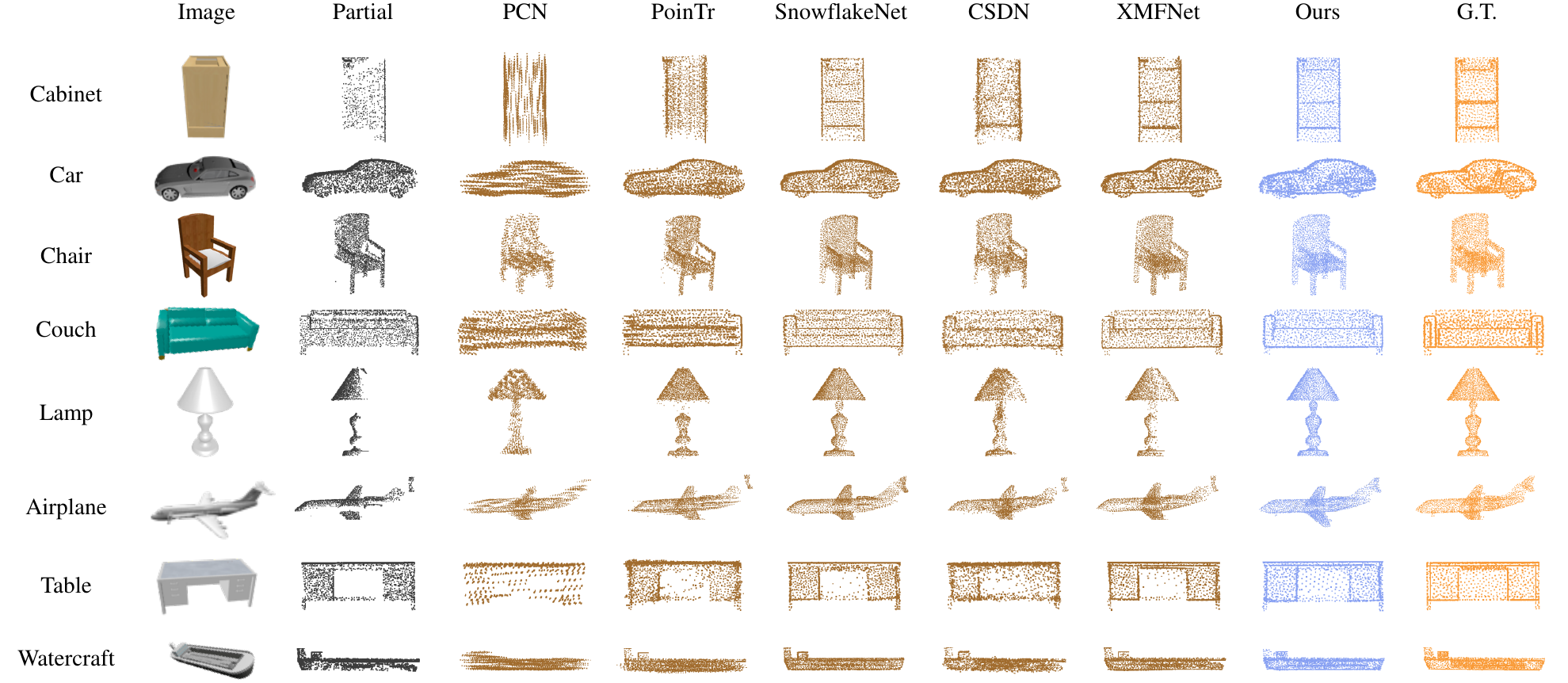}
   \caption{Qualitative comparisons to the state-of-the-art methods on all eight categories of the ShapeNet-ViPC dataset.}
   \label{fig:shapenet}
\end{figure*}

\begin{figure}[t]
  \centering
   \includegraphics[width=1\linewidth]{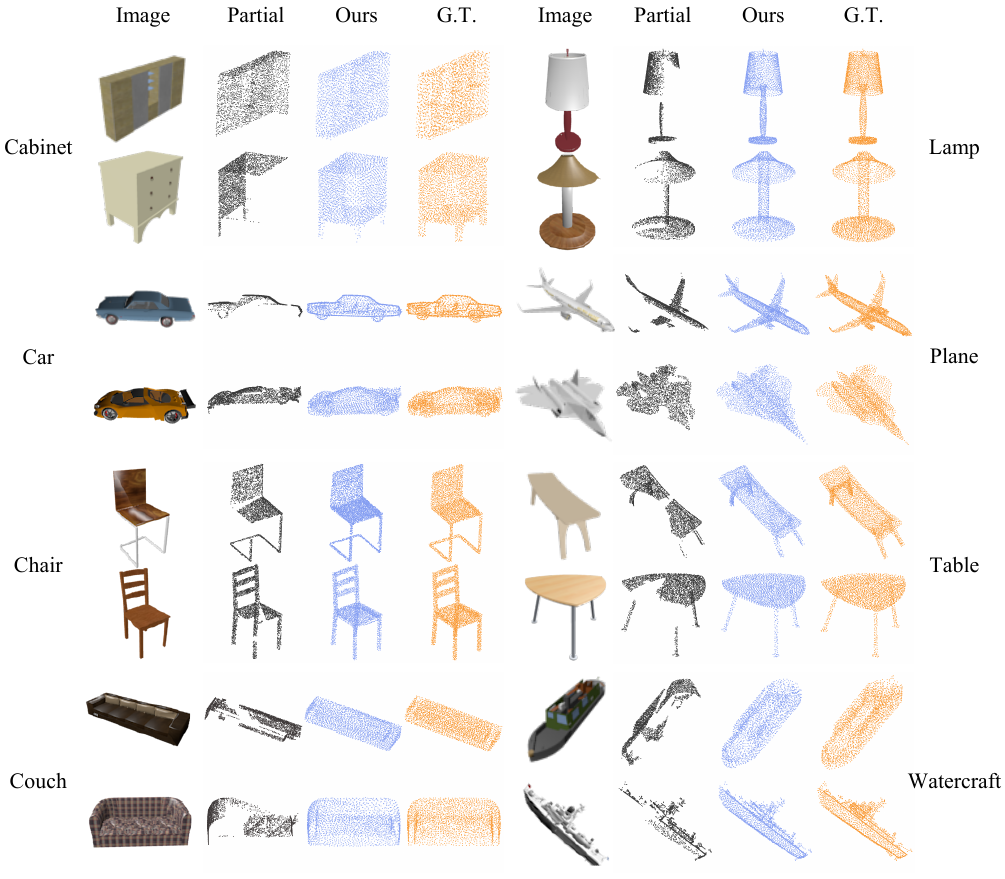}
   \caption{More qualitative results of the DuInNet on all eight categories of the ShapeNet-ViPC dataset.}
   \vspace{-4ex}
   \label{fig:shapenetvipc_others}
\end{figure}

We adopt the PyTorch 2.0.1 framework~\cite{pytorch} and train models on 4 NVIDIA 3090 GPUs with distributed data parallel mode.
In the ShapeNet-ViPC dataset, we adopt the same experimental settings as XMFNet~\cite{xmfnet}. 
In the ModelNet-MPC dataset, we train all models for 120 epochs with a batch size of 128. The learning rate is initialized to $1\times10^{-4}$ and decayed by 0.1 at epochs 25 and 75.

\subsection{Results on the ShapeNet-ViPC Dataset}
To demonstrate the superiority of our method, in this section, we compare our DuInNet with other single-modal and multimodal methods on the ShapeNet-ViPC dataset.
\subsubsection{Quantitative results}
To quantitatively evaluate the performance of different methods, we report all 8 categories and their mean metrics in Table~\ref{tab:shapenetvipc}.
We can observe that our DuInNet achieves a significant performance improvement in all 8 categories over the state-of-the-art single-modal methods.
Moreover, compared with the state-of-the-art multimodal methods, DuInNet also achieves the best performance. It significantly reduces mean CD by 0.296 ($\downarrow$\;20.5\%) and {increases mean FS by 0.055 ($\uparrow$\;6.9\%)}, compared with XMFNet~\cite{xmfnet}.
{It is worth noting that the performance of XMFNet~\cite{xmfnet} is not even as good as that of single-modal method SnowflakeNet~\cite{snowflakenet}, which only uses partial point clouds for the shape completion.
However, our DuInNet achieves better completion performance, even though it has a simpler point cloud generation network than SnowflakeNet~\cite{snowflakenet}.}
This is because the unidirectional fusion structure in XMFNet ignores the shape prior contained in the image modality, while our DuInNet makes deep interaction between dual modalities and directly restores the point clouds from image modality.

\subsubsection{Qualitative results}
We also visually compare the completion results of different methods. As shown in Fig.~\ref{fig:shapenet}, our DuInNet can effectively restore the detailed structures of objects based on texture characteristic provided by images.
{
We summarize the advantages of our method into three types.
Firstly, DuInNet can generate smooth and complete surfaces without many outliers.
For example, compared with PCN~\cite{pcn}, PoinTr~\cite{pointr} and CSDN~\cite{csdn}, our DuInNet can generate sharp and flat object contours with fewer noise points.
Secondly, DuInNet excels at extracting features from existing parts of objects and generating overall features with detailed shapes.
For example, compared with XMFNet~\cite{xmfnet}, our DuInNet can generate precise shapes of the missing cover and pillar of the lamp according to existing point clouds on the other side, and generate smooth surface and clear shape boundary with the assistance of input images.
Thirdly, DuInNet can capture and generate more realistic and semantic object structures.
For example, compared with SnowflakeNet~\cite{snowflakenet}, our DuInNet can generate finer object shapes like the back panel of the table in Fig.~\ref{fig:shapenet}.
Meanwhile, we give more qualitative results of the DuInNet on all 8 categories of the ShapeNet-ViPC dataset in Fig.~\ref{fig:shapenetvipc_others}.
It shows that DuInNet has good completion performance for objects of different shapes.}
In summary, by deeply interacting features of two modalities and fully leveraging the shape priors contained in these two modalities, our DuInNet performs remarkably well on the ShapeNet-ViPC dataset.

\begin{table*}[t]
\caption{Quantitative comparisons to the state-of-the-art methods on the proposed ModelNet-MPC dataset for the \textbf{fully supervised} completion task in terms of Mean Chamfer Distance per point (CD, $\times 10^{-3}$) and Mean F-Score @ 0.001 (FS). Lower CD and higher FS are better. The best results are highlighted in \textbf{boldface}.}
\centering
\renewcommand\arraystretch{1.2}
\resizebox{\textwidth}{!}{
\begin{tabular}{c|cc|cc|cc|cc|cc|cc|cc|cc|cc|cc|cc}
\hline  Methods
            & \multicolumn{14}{c|}{Single-modal}                                                 & \multicolumn{8}{c}{Multi-modal}                                       \\ \hline 
\multirow{-4}{*}{Categories} & \multicolumn{2}{c|}{\rotatebox{45}{FoldingNet~\cite{foldingnet}}} & \multicolumn{2}{c|}{\rotatebox{45}{PCN~\cite{pcn}} }  & \multicolumn{2}{c|}{\rotatebox{45}{TopNet~\cite{topnet}} }& \multicolumn{2}{c|}{\rotatebox{45}{PoinTr~\cite{pointr}}} &\multicolumn{2}{c|}{ \rotatebox{45}{SnowflakeNet~\cite{snowflakenet}} }  & \multicolumn{2}{c|}{\rotatebox{45}{AdaPoinTr~\cite{adapointr}}} & \multicolumn{2}{c|}{\rotatebox{45}{Seedformer~\cite{seedformer}}} & \multicolumn{2}{c|}{\rotatebox{45}{ViPC~\cite{vipc}} }& \multicolumn{2}{c|}{\rotatebox{45}{CSDN~\cite{csdn}}}  & \multicolumn{2}{c|}{\rotatebox{45}{XMFNet~\cite{xmfnet}}} & \multicolumn{2}{c}{\cellcolor[HTML]{EFEFEF}\textbf{\rotatebox{45}{DuInNet (Ours)}}}  \\
\multicolumn{1}{c|}{} &
  CD &
  \multicolumn{1}{c|}{FS} &
  CD &
  \multicolumn{1}{c|}{FS} &
  CD &
  \multicolumn{1}{c|}{FS} &
  CD &
  \multicolumn{1}{c|}{FS} &
  CD &
  \multicolumn{1}{c|}{FS} &
  CD &
  \multicolumn{1}{c|}{FS} &
  CD &
  \multicolumn{1}{c|}{FS} &
  CD &
  \multicolumn{1}{c|}{FS} &
  CD &
  \multicolumn{1}{c|}{FS} &
  CD &
  \multicolumn{1}{c|}{FS} &
  \cellcolor[HTML]{EFEFEF}\textbf{CD} &
  \cellcolor[HTML]{EFEFEF}\textbf{FS} \\ \hline\hline
airplane &
  2.320 & 0.749 &
  1.694 & 0.839 &
  1.532 & 0.839 &
  0.992 & 0.909 &
  0.788 &0.928          &
  0.773 &\textbf{0.937} &
  1.058 &0.883 &
  2.397 &0.739 &
  1.548 &0.808 &
  1.011 &0.888 &
  \cellcolor[HTML]{EFEFEF}\textbf{0.732} & \cellcolor[HTML]{EFEFEF}0.935\\
bathtub &
  5.241 &0.443 &
  5.219 &0.469 &
  4.837 & 0.443 &
  4.034 &0.583 &
  2.993 &\textbf{0.645} &
  3.008 &0.622          &
  3.229 &0.617 & 
  6.641 &0.430 &
  4.987 &0.520 &
  3.371 &0.577 &
  \cellcolor[HTML]{EFEFEF}\textbf{2.938} &\cellcolor[HTML]{EFEFEF}0.644  \\
bed &
  3.988 &0.452 &
  3.610 & 0.504 &
  3.560 &0.487 &
  2.585 &0.636 &
  2.321 & 0.646          &
  2.352 &0.633          &
  2.408 &0.625 & 
  5.221 &0.445 &
  3.914 &0.532 &
  2.714 &0.588 &
  \cellcolor[HTML]{EFEFEF}\textbf{2.297}& \cellcolor[HTML]{EFEFEF}\textbf{0.657} \\
bench &
  5.025 &0.541 &
  4.961 &0.609 &
  4.381 &0.595 &
  2.765 &0.746 & 
  2.356 &0.775          &
  2.363 &0.765          & 
  2.347 &0.754 &
  6.937 & 0.567 &
  5.325 &0.647 &
  2.549 &0.711 &
  \cellcolor[HTML]{EFEFEF}\textbf{2.031}  & \cellcolor[HTML]{EFEFEF}\textbf{0.778}\\
bookshelf &
  5.023 &0.432 &
  4.429 &0.503 &
  4.527 &0.465 &
  3.006 &0.619 &
  2.410 &\textbf{0.696} &
  2.446 &0.663          &
  \textbf{2.361} &0.676 & 
  5.747 &0.450 & 
  4.665 &0.533 &
  3.196 &0.564 &
  \cellcolor[HTML]{EFEFEF}2.726 &\cellcolor[HTML]{EFEFEF}0.653\\
bottle &
  2.493 &0.732 &
  2.043 & 0.799 & 
  1.998 &0.786 &
  1.582 &0.838 & 
  1.372 &0.864          &
  1.406 &0.864          & 
  1.494 &0.833 &
  3.024 &0.670 &
  2.497 &0.759 &
  1.560 & 0.822 &
  \cellcolor[HTML]{EFEFEF}\textbf{1.352} &\cellcolor[HTML]{EFEFEF}\textbf{0.887}\\
bowl &
  7.623 &0.331 &
  7.330 &0.365 &
  7.324 &0.345 &
  6.178 &0.496 &
  5.157 &\textbf{0.565} & 
  \textbf{4.871} &0.501          &
  5.742 &0.547 & 
  13.083 &0.353 &
  13.603 &0.453 &
  5.304 &0.523 &
  \cellcolor[HTML]{EFEFEF}5.145 & \cellcolor[HTML]{EFEFEF}0.529 \\
car &
  4.182 &0.442 &
  3.662 &0.494 &
  3.872 &0.456 &
  3.077 &0.563 &
  \textbf{2.615} &0.592&          
  2.672 &0.569     &     
  2.930 & 0.549 &
  5.079 & 0.434 &
  4.014 &       0.508  &
  3.051 &       0.543 &
  \cellcolor[HTML]{EFEFEF}2.673&\cellcolor[HTML]{EFEFEF}\textbf{0.621} \\
chair &
  5.416 &0.470 &
  4.728 &0.558 &
  4.265 &0.541 &
  3.030 &0.690 &
  2.453 &\textbf{0.739}&
  2.362 &0.736    &      
  2.598 &0.699 &
  7.330 &0.535 &
  4.853 &0.611 &
  3.004 &0.659 &
  \cellcolor[HTML]{EFEFEF}\textbf{2.343} &\cellcolor[HTML]{EFEFEF}0.730    \\
cone &
  3.440 &0.618 &
  3.210 &0.696 &
  3.097 &0.658 &
  2.284 &0.786 &
  2.049 &0.819       &   
  2.089 &0.823    &      
  2.250 &0.774 &
  4.907 &0.575 &
  4.133 &0.703 &
  2.197 &0.715 &
  \cellcolor[HTML]{EFEFEF}\textbf{1.808} & \cellcolor[HTML]{EFEFEF}\textbf{0.853}\\
cup &
  8.337 &0.315 &
  7.854 &0.346 &
  7.618 &0.315 &
  5.785 &0.448 &
  5.314 &0.504       &   
  4.961 &0.474    &      
  5.606 &0.491 &
  9.037 &0.357 &
  7.257 &0.420 &
  5.288 &0.474 &
  \cellcolor[HTML]{EFEFEF}\textbf{4.537}& \cellcolor[HTML]{EFEFEF}\textbf{0.518} \\
curtain &
  2.316 &0.731 &
  2.279 &0.786 &
  2.022 &0.770 &
  1.559 &0.786 &
  1.061 &0.883       &   
  1.167 &0.865    &      
  1.077 &0.862 &
  3.425 &0.688 &
  3.408 &0.765 &
  1.283 &0.838 &
  \cellcolor[HTML]{EFEFEF}\textbf{1.017} & \cellcolor[HTML]{EFEFEF}\textbf{0.888} \\
desk &
  6.918 &0.412 &
  6.749 &0.461 &
  6.082 &0.446 &
  3.952 &0.632 &
  3.316 &\textbf{0.684}& 
  3.333 &0.655    &      
  3.334 &0.661 &
  8.908 &0.458 &
  6.695 &0.537 &
  4.003 &0.588 &
  \cellcolor[HTML]{EFEFEF}\textbf{3.312}& \cellcolor[HTML]{EFEFEF}0.667    \\
door &
  2.332 &0.749 &
  1.907 &0.844 &
  1.968 &0.804 &
  1.655 &0.764 &
  1.092 &0.874       &   
  1.187 &0.878    &      
  1.185 &0.843 &
  2.784 &0.735 &
  2.036 &0.798 &
  1.038 &0.803 &
  \cellcolor[HTML]{EFEFEF}\textbf{0.964}& \cellcolor[HTML]{EFEFEF}\textbf{0.902} \\
dresser &
  4.139 &0.449 &
  3.357 &0.513 &
  3.670 &0.457 &
  2.841 &0.576 &
  2.483 &0.615       &   
  2.486 &0.577    &      
  2.640 &0.592 &
  5.531 &0.397 &
  4.697 &0.492 &
  2.869 &0.575 &
  \cellcolor[HTML]{EFEFEF}\textbf{2.416}& \cellcolor[HTML]{EFEFEF}\textbf{0.632} \\
flower\_pot &
  9.791 &0.283 &
  8.742 &0.326 &
  8.115 &0.303 &
  5.754 &0.460 &
  5.239 &0.500 &         
  4.990 &0.480    &      
  5.155 &0.474 &
  8.817 &0.372 &
  7.443 &0.418 &
  5.301 &0.476 &
  \cellcolor[HTML]{EFEFEF}\textbf{4.584} & \cellcolor[HTML]{EFEFEF}\textbf{0.511}\\
glass\_box &
  4.246 &0.421 &
  3.212 &0.501 &
  3.521 &0.456 &
  3.230 &0.557 &
  2.392 &0.595       &   
  2.485 &0.554    &      
  3.155 &0.552 &
  4.815 &0.401 &
  4.307 &0.491 &
  2.751 &0.507 &
  \cellcolor[HTML]{EFEFEF}\textbf{2.238} & \cellcolor[HTML]{EFEFEF}\textbf{0.622} \\
guitar &
  1.326 &0.876 &
  1.709 &0.880 &
  1.063 &0.911 &
  0.501 &0.961 &
  0.400 &0.978          &
  0.406 & \textbf{0.980}& 
  0.361 &0.976 &
  1.266 &0.878 &
  1.494 &0.892 &
  0.506 &0.955 &
  \cellcolor[HTML]{EFEFEF}\textbf{0.336} & \cellcolor[HTML]{EFEFEF}0.979     \\
keyboard &
  1.462 &0.805 &
  1.245 &0.864 &
  1.258 &0.839 &
  0.831 &0.905 &
  0.766 &0.904       &   
  0.835 &0.899    &      
  0.784 &0.889 &
  1.876 &0.767 &
  2.003 &0.792 &
  0.824 &0.857 &
  \cellcolor[HTML]{EFEFEF}\textbf{0.734}& \cellcolor[HTML]{EFEFEF}\textbf{0.937} \\
lamp &
  15.680 &0.408 &
  18.640 &0.460 &
  16.043 &0.465 &
  8.876 &0.658 &
  6.111 &0.758       &   
  6.791 &0.760    &      
  4.623 &0.749 &
  19.236 &0.488 &
  10.592 &0.603 &
  4.986 &0.667 &
  \cellcolor[HTML]{EFEFEF}\textbf{3.208}& \cellcolor[HTML]{EFEFEF}\textbf{0.760}  \\
laptop &
  3.211 &0.596 &
  2.602 &0.710 &
  2.483 &0.648 &
  1.560 &0.773 &
  1.364 &0.790       &   
  1.369 &0.786    &      
  1.682 &0.740 &
  4.588 &0.577 &
  3.042 &0.672 &
  1.531 &0.711 &
  \cellcolor[HTML]{EFEFEF}\textbf{1.177} & \cellcolor[HTML]{EFEFEF}\textbf{0.823}\\
mantel &
  4.346 &0.425 &
  2.959 &0.558 &
  3.156 &0.514 &
  2.450 &0.638 &
  \textbf{1.792} & \textbf{0.728}&
  1.904 &0.693    &      
  2.197 &0.692 &
  5.857 &0.443 &
  3.905 &0.545 &
  2.327 &0.644 &
  \cellcolor[HTML]{EFEFEF}1.982 & \cellcolor[HTML]{EFEFEF}0.680 \\
monitor &
  4.265 &0.525 &
  3.465 &0.581 &
  3.506 &0.547 &
  2.794 &0.635 &
  2.250 &0.693       &   
  2.201 &0.676    &      
  2.464 &0.667 &
  4.858 &0.508 &
  3.795 &0.579 &
  2.489 &0.637 &
  \cellcolor[HTML]{EFEFEF}\textbf{2.159} & \cellcolor[HTML]{EFEFEF}\textbf{0.708}\\
night\_stand &
  5.172 &0.389    &  
  4.695 &0.444 &
  4.730 &0.400  &
  3.365 &0.557  &
  \textbf{2.888} & \textbf{0.596}& 
  2.891 &0.556     &
  3.049 &0.576    &  
  7.198 &0.390 &
  5.583 &0.471 &
  3.666 &0.543  &
  \cellcolor[HTML]{EFEFEF}2.916 & \cellcolor[HTML]{EFEFEF}0.594 \\
person &
  6.593 &0.421 &
  7.071 &0.526 &
  5.486 &0.511 &
  3.245 &0.670 &
  2.685 &0.741       &   
  2.513 &0.743    &      
  2.364 &0.720 &
  5.265 &0.600 &
  5.725 &0.642 &
  2.541 &0.707 &
  \cellcolor[HTML]{EFEFEF}\textbf{1.977} & \cellcolor[HTML]{EFEFEF}\textbf{0.766} \\
piano &
  7.728 &0.349 &
  6.525 &0.420 &
  6.173 &0.382 &
  4.502 &0.545 &
  3.687 &0.607       &   
  3.605 &0.577    &      
  4.072 &0.568 &
  7.421 &0.399 &
  5.716 &0.480 &
  3.763 &0.526 &
  \cellcolor[HTML]{EFEFEF}\textbf{3.116} & \cellcolor[HTML]{EFEFEF}\textbf{0.616} \\
plant &
  12.860 &0.260 &
  12.622 &0.296 &
  9.721 &0.305 &
  6.485 &0.485 &
  5.908 &0.525       &   
  5.978 &0.505    &      
  5.157 &0.512 &
  9.170 &0.429 &
  8.590 &0.457 &
  5.929 &0.516 &
  \cellcolor[HTML]{EFEFEF}\textbf{4.828}& \cellcolor[HTML]{EFEFEF}\textbf{0.546} \\
radio &
  5.644 &0.395 &
  5.692 &0.459 &
  4.904 &0.440 &
  3.068 &0.641 &
  2.698 &0.691       &   
  2.692 &0.658    &      
  2.644 &0.672 &
  7.513 &0.431 &
  6.005 &0.535 &
  2.983 &0.613 &
  \cellcolor[HTML]{EFEFEF}\textbf{2.491} & \cellcolor[HTML]{EFEFEF}\textbf{0.698}\\
range\_hood &
  5.644 &0.386 &
  4.663 &0.459 &
  4.186 &0.437 &
  3.541 &0.560 &
  \textbf{2.451} &\textbf{0.667}& 
  2.453 &0.646    &      
  3.029 &0.633 &
  7.054 &0.396 &
  5.014 &0.494 &
  3.086 &0.578 &
  \cellcolor[HTML]{EFEFEF}2.561& \cellcolor[HTML]{EFEFEF}0.621   \\
sink &
  6.577 &0.440 &
  6.488 &0.517 &
  5.656 &0.477 &
  3.805 &0.671 &
  3.036 &\textbf{0.738} &
  2.927 &0.724    &      
  3.087 &0.710 &
  8.737 &0.461 &
  6.056 &0.575 &
  3.512 &0.644 &
  \cellcolor[HTML]{EFEFEF}\textbf{2.694} & \cellcolor[HTML]{EFEFEF}0.730 \\
sofa &
  3.848 &0.449 &
  3.399 &0.499 &
  3.477 &0.480 &
  2.790 &0.586 &
  \textbf{2.316} &0.623&          
  2.365 &0.606    &      
  2.476 &0.592 &
  5.155 &0.424 &
  4.118 &0.495 &
  2.814 &0.554 &
  \cellcolor[HTML]{EFEFEF}2.449 & \cellcolor[HTML]{EFEFEF}\textbf{0.625} \\
stairs &
  12.590 &0.334 &
  17.354 &0.333 &
  11.608 &0.362 &
  7.065 &0.590 &
  5.877 &0.665       &   
  5.875 &0.642    &      
  4.562 &0.673 &
  10.089 &0.512 &
  10.572 &0.563 &
  4.625 &0.634 &
  \cellcolor[HTML]{EFEFEF}\textbf{3.503}& \cellcolor[HTML]{EFEFEF}\textbf{0.703} \\
stool &
  7.896 &0.409 &
  10.734 &0.440 &
  8.482 &0.455 &
  4.359 &0.674 &
  3.635 &\textbf{0.738}& 
  3.590 &0.736    &      
  3.074 &0.724 &
  11.570 &0.502 &
  6.926 &0.598 &
  4.464 &0.632 &
  \cellcolor[HTML]{EFEFEF}\textbf{2.963}  & \cellcolor[HTML]{EFEFEF}0.729   \\
table &
  4.115 &0.569 &
  3.831 &0.682 &
  3.197 &0.684 &
  2.101 &0.818 &
  1.914 &0.827       &   
  2.006 &0.816    &      
  2.011 &0.798 &
  5.743 &0.610 &
  3.696 &0.694 &
  2.107 &0.759 &
  \cellcolor[HTML]{EFEFEF}\textbf{1.664}& \cellcolor[HTML]{EFEFEF}\textbf{0.832} \\
tent &
  7.254 &0.367 &
  7.788 &0.392 &
  7.160 &0.366 &
  4.169 &0.600 &
  3.617 &0.666       &   
  3.865 &0.623    &      
  3.676 &0.641 &
  9.960 &0.407 &
  7.350 &0.518 &
  3.971 &0.593 &
  \cellcolor[HTML]{EFEFEF}\textbf{3.092} & \cellcolor[HTML]{EFEFEF}\textbf{0.676} \\
toilet &
  5.770 &0.387 &
  4.885 &0.435 &
  4.884 &0.407 &
  4.021 &0.527 &
  \textbf{3.145} &\textbf{0.590} &
  3.213 &0.559    &      
  3.435 &0.547 &
  7.805 &0.391 &
  5.725 &0.464 &
  3.826 &0.504 &
  \cellcolor[HTML]{EFEFEF}3.279 & \cellcolor[HTML]{EFEFEF}0.572 \\
tv\_stand &
  5.911 &0.385 &
  5.472 &0.459 &
  5.487 &0.421 &
  3.908 &0.558 &
  3.408 &0.605       &   
  3.439 &0.572    &      
  3.311 &0.591 &
  7.029 &0.402 &
  5.876 &0.484 &
  3.804 &0.547 &
  \cellcolor[HTML]{EFEFEF}\textbf{3.255} & \cellcolor[HTML]{EFEFEF}\textbf{0.612}\\
vase &
  6.041 &0.441 &
  5.219 &0.486 &
  5.117 &0.467 &
  3.952 &0.588 &
  3.408 &0.639          &
  \textbf{3.366} &0.623&
  3.644 &0.610 &
  8.058 &0.434 &
  6.522 &0.516 &
  3.906 &0.585 &
  \cellcolor[HTML]{EFEFEF}3.397 & \cellcolor[HTML]{EFEFEF}\textbf{0.640} \\
wardrobe &
  4.335 &0.498 &
  3.834 &0.574 &
  4.051 &0.518 &
  2.694 &0.662 &
  2.421 &0.700       &   
  2.516 &0.659    &      
  2.300 &0.678 &
  5.124 &0.438 &
  4.131 &0.552 &
  2.543 &0.640 &
  \cellcolor[HTML]{EFEFEF}\textbf{2.237}& \cellcolor[HTML]{EFEFEF}\textbf{0.725} \\
xbox &
  4.436 &0.503 &
  4.078 &0.565 &
  4.037 &0.507 &
  2.830 &0.644 &
  2.475 &0.687       &   
  2.571 &0.642    &      
  2.393 &0.664 &
  5.407 &0.432 &
  4.235 &0.555 &
  2.503 &0.630 &
  \cellcolor[HTML]{EFEFEF}\textbf{2.018} & \cellcolor[HTML]{EFEFEF}\textbf{0.722} \\ \hline
Mean &
  5.152 &0.488 &
  4.751 &0.554 &
  4.395 &0.531 &
  3.127 &0.659 &
  2.596 &0.705       &   
  2.612 &0.687    &      
  2.691 &0.676 &
  6.204 &0.501 &
  4.782 &0.580 &
  2.937 &0.644 &
  \cellcolor[HTML]{EFEFEF}\textbf{2.428} & \cellcolor[HTML]{EFEFEF}\textbf{0.707} \\ \hline
\end{tabular}
}
\label{tab:mpc1}
\end{table*}

\subsection{Results on the ModelNet-MPC Dataset}
\begin{figure*}[!p]
  \centering
  \vspace{-2ex}
   \includegraphics[width=1\linewidth]{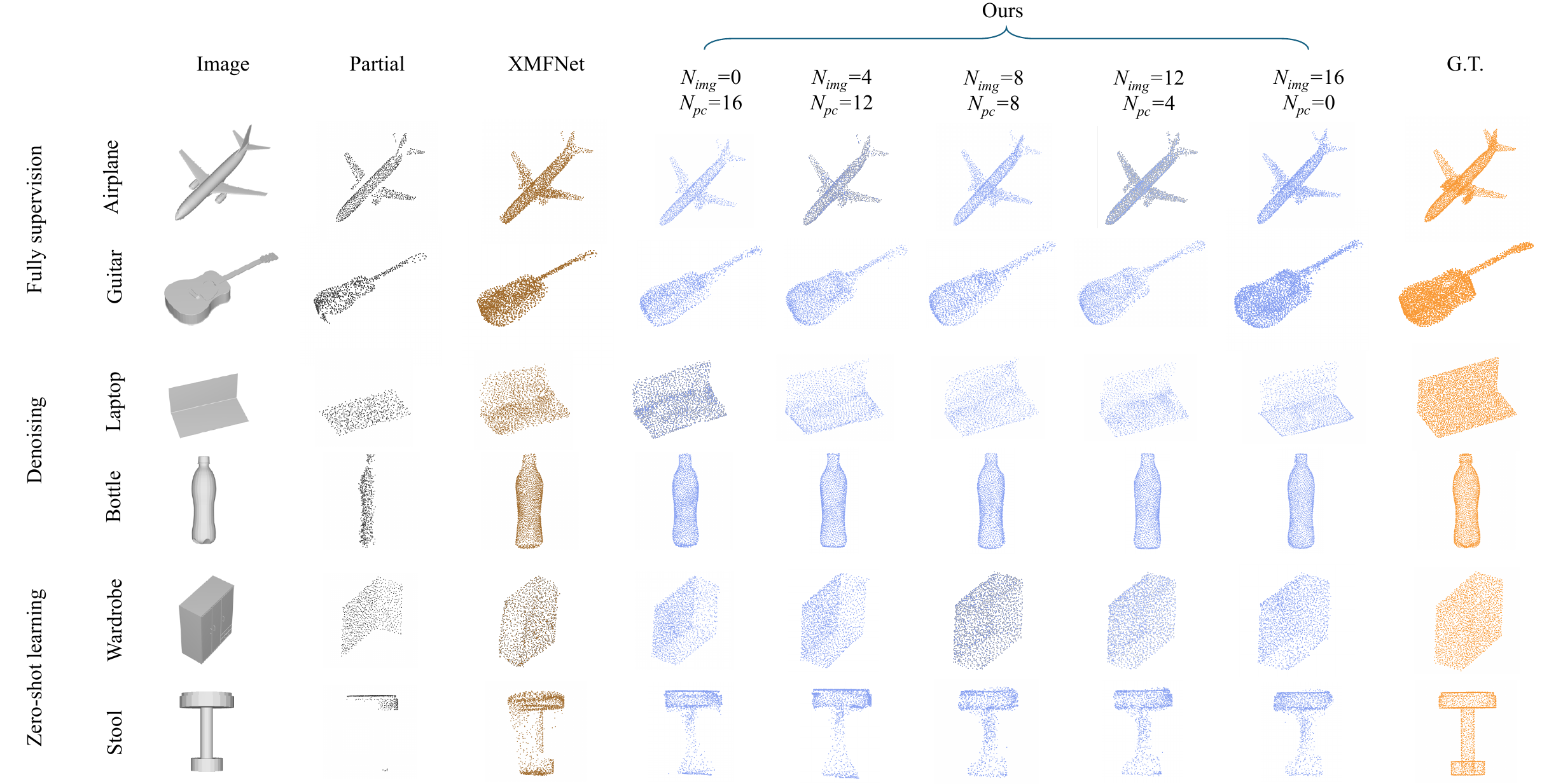}
   \caption{Qualitative comparisons of completion performance between XMFNet and five variants of DuInNet (different modalities weights for the adaptive point generator) on three expanded completion tasks of the ModelNet-MPC dataset.}
   \label{fig:modelnet}
\end{figure*}

In this section, we comprehensively compare our DuInNet with other methods on ModelNet-MPC for three different tasks: fully supervised completion, denoising completion and zero-shot learning completion {in Table~\ref{tab:mpc1}, Table~\ref{tab:mpc2} and Table~\ref{tab:mpc3}, respectively.
Moreover, we visually compare XMFNet~\cite{xmfnet} and five variants of DuInNet (different modalities weights for the adaptive point generator) on the ModelNet-MPC dataset in Fig.~\ref{fig:modelnet}.
They not only verify the applicability of the proposed ModelNet-MPC benchmark, but also demonstrate the superiority, noise robustness, category generalization and transfer ability of the DuInNet.}
\begin{table*}[!p]
\caption{Quantitative comparisons to the state-of-the-art methods on the proposed ModelNet-MPC dataset for the \textbf{denoising} point cloud completion task in terms of Mean Chamfer Distance per point (CD, $\times 10^{-3}$) and Mean F-Score @ 0.001 (FS). Lower CD and higher FS are better. The best results are highlighted in \textbf{boldface}.}
\centering
\renewcommand\arraystretch{1.1}
\resizebox{\textwidth}{!}{
\begin{tabular}{c|cc|cc|cc|cc|cc|cc|cc|cc|cc|cc}
\hline  Methods
            & \multicolumn{12}{c|}{Single-modal}                                                 & \multicolumn{8}{c}{Multi-modal}                                        \\ \hline 
\multirow{-2.5}{*}{Categories} & \multicolumn{2}{c|}{\rotatebox{30}{FoldingNet~\cite{foldingnet}}} & \multicolumn{2}{c|}{\rotatebox{30}{PCN~\cite{pcn}}}   &\multicolumn{2}{c|}{ \rotatebox{30}{TopNet~\cite{topnet}}} & \multicolumn{2}{c|}{\rotatebox{30}{PoinTr~\cite{pointr}}} & \multicolumn{2}{c|}{\rotatebox{30}{AdaPoinTr~\cite{adapointr}}} & \multicolumn{2}{c|}{\rotatebox{30}{Seedformer~\cite{seedformer}}} & \multicolumn{2}{c|}{\rotatebox{30}{ViPC~\cite{vipc}}} & \multicolumn{2}{c|}{\rotatebox{30}{CSDN~\cite{csdn}}}  & \multicolumn{2}{c|}{\rotatebox{30}{XMFNet~\cite{xmfnet}}} & \multicolumn{2}{c}{\cellcolor[HTML]{EFEFEF}\textbf{\rotatebox{30}{DuInNet (Ours)}}   \cellcolor[HTML]{EFEFEF} }\\ 

\multicolumn{1}{c|}{} &
  CD &
  \multicolumn{1}{c|}{FS} &
  CD &
  \multicolumn{1}{c|}{FS} &
  CD &
  \multicolumn{1}{c|}{FS} &
  CD &
  \multicolumn{1}{c|}{FS} &
  CD &
  \multicolumn{1}{c|}{FS} &
  CD &
  \multicolumn{1}{c|}{FS} &
  CD &
  \multicolumn{1}{c|}{FS} &
  CD &
  \multicolumn{1}{c|}{FS} &
  CD &
  \multicolumn{1}{c|}{FS} &
  \cellcolor[HTML]{EFEFEF}\textbf{CD} &
  \cellcolor[HTML]{EFEFEF}\textbf{FS} \\ \hline\hline

airplane     & 2.153  & 0.797& 4.437     & 0.828     & 1.969  & 0.797 & 1.381      & \textbf{0.912}    & 1.479      & 0.891     & 1.644  & 0.881& 3.461 & 0.656   & 1.956   & 0.804 & 1.621 & 0.837 & \textbf{1.142} \cellcolor[HTML]{EFEFEF} & 0.893          \cellcolor[HTML]{EFEFEF} \\
bathtub      & 4.752& 0.471  & 4.878  & 0.383        & 5.139 & 0.409 & 3.886      & 0.575      & 3.724  & 0.577         & 4.443 & 0.515 & 7.639   & 0.363 & 6.899    & 0.450& 4.575 & 0.466 & \textbf{3.475} \cellcolor[HTML]{EFEFEF} & \textbf{0.578} \cellcolor[HTML]{EFEFEF} \\
bed          & 3.793 & 0.491 & 6.323    & 0.422      & 3.992  & 0.444& 2.995   & 0.589         & 3.025     & \textbf{0.591}      & 3.617 & 0.504 & 8.008   & 0.380 & 5.806    & 0.478& 3.730 & 0.488& \textbf{2.952} \cellcolor[HTML]{EFEFEF} & 0.578          \cellcolor[HTML]{EFEFEF}  \\
bench        & 4.432 & 0.588 & 4.459   & 0.565      & 4.406  & 0.569 & 3.102      & 0.714    & 2.717& \textbf{0.736}          & 3.212 & 0.690  & 8.395& 0.490    & 6.408 & 0.578    & 3.721& 0.617  & \textbf{2.602} \cellcolor[HTML]{EFEFEF} & 0.719          \cellcolor[HTML]{EFEFEF} \\
bookshelf    & 4.525& 0.474  & 4.453 & 0.434         & 4.765 & 0.434  & 3.233   & 0.606        & \textbf{2.929}  & \textbf{0.620}& 3.413& 0.567  & 6.779 & 0.382   & 6.106 & 0.462    & 4.557 & 0.472& 3.284          \cellcolor[HTML]{EFEFEF} & 0.577          \cellcolor[HTML]{EFEFEF} \\
bottle       & 2.463  & 0.761 & 5.391   & 0.592        & 2.741 & 0.707 & 1.982        & 0.849   & 2.436 & 0.809              & 2.331  & 0.789 & 6.677  & 0.567  & 4.403 & 0.728    & 2.200 & 0.787 & \textbf{1.683} \cellcolor[HTML]{EFEFEF} & \textbf{0.852} \cellcolor[HTML]{EFEFEF} \\
bowl         & 22.109& 0.349 & \textbf{5.023}& 0.257  & 23.047 & 0.290 & 17.648     & 0.415     & 28.097        & 0.412  & 12.601& 0.365 & 3979 & 0.237 & 3616 & 0.318& 9.057 & 0.332& 11.544         \cellcolor[HTML]{EFEFEF} & \textbf{0.416} \cellcolor[HTML]{EFEFEF} \\
car          & 3.919& 0.470  & 7.782 & 0.432          & 4.412  & 0.420& \textbf{3.236}& 0.552  & 3.359    & 0.528       & 3.957    & 0.462& 6.343  & 0.362  & 4.761& 0.456    & 3.941 & 0.475& 3.295          \cellcolor[HTML]{EFEFEF} & \textbf{0.555} \cellcolor[HTML]{EFEFEF} \\
chair        & 4.751 & 0.511 & 7.649    & 0.512      & 4.594   & 0.508& 3.248   & 0.674        & \textbf{2.993}& \textbf{0.690} & 3.596 & 0.640   & 8.280  & 0.469   & 6.074 & 0.557   & 4.420 & 0.544& 3.030          \cellcolor[HTML]{EFEFEF} & 0.668          \cellcolor[HTML]{EFEFEF} \\
cone         & 3.334  & 0.699& \textbf{2.050}& 0.556 & 3.841 & 0.581  & 3.525    & 0.798       & 3.550     & 0.751        & 3.486 & 0.722  & 8.035& 0.468    & 5.817   & 0.665 & 3.414& 0.712  & 2.496          \cellcolor[HTML]{EFEFEF} & \textbf{0.804} \cellcolor[HTML]{EFEFEF} \\
cup          & 11.244 & 0.312& \textbf{4.513}  & 0.233& 12.984 & 0.274& 11.246  & 0.367         & 11.002 & 0.381            & 12.697& 0.311 & 71.301   & 0.253& 35.666 & 0.320   & 8.479& 0.331 & 6.336          \cellcolor[HTML]{EFEFEF} & \textbf{0.411} \cellcolor[HTML]{EFEFEF} \\
curtain      & 5.557& 0.748   & 4.047   & 0.744       & 6.615   & 0.734& 3.602  & 0.825         & 8.217    & 0.838        & 3.823 & 0.820  & 79.288 & 0.610  & 67.762 & 0.710  & 2.618& 0.767 & \textbf{2.491} \cellcolor[HTML]{EFEFEF} & \textbf{0.842} \cellcolor[HTML]{EFEFEF} \\
desk         & 6.215 & 0.445 & \textbf{3.545} & 0.404& 6.185 & 0.431  & 4.307       & 0.598    & 3.828 & \textbf{0.639}          & 4.257  & 0.587 & 10.024 & 0.395  & 8.552  & 0.472   & 5.781 & 0.475 & 3.987         \cellcolor[HTML]{EFEFEF} & 0.601          \cellcolor[HTML]{EFEFEF} \\
door         & 2.351& 0.804  & 4.014  & 0.797         & 2.485  & 0.758 & 1.814  & 0.866        & 2.450 & 0.851         & 1.944  & 0.836& 9.344   & 0.671 & 3.740   & 0.777 & 1.739 & 0.808& \textbf{1.438} \cellcolor[HTML]{EFEFEF} & \textbf{0.870} \cellcolor[HTML]{EFEFEF} \\
dresser      & 3.676 & 0.498 & 6.591    & 0.350       & 4.005 & 0.425 & \textbf{2.894}& \textbf{0.565}  & 3.006    & 0.552       & 3.613   & 0.454 & 6.616  & 0.310   & 5.519 & 0.463   & 3.942& 0.484 & 3.081          \cellcolor[HTML]{EFEFEF} & 0.556          \cellcolor[HTML]{EFEFEF} \\
flower\_pot  & 8.267& 0.308   & 5.605    & 0.271       & 8.147& 0.288  & 6.437      & 0.414      & 5.839   & 0.427       & 7.083 & 0.357  & 10.393  & 0.294  & 9.183 & 0.352   & 7.066 & 0.355& \textbf{5.285} \cellcolor[HTML]{EFEFEF} & \textbf{0.444} \cellcolor[HTML]{EFEFEF} \\
glass\_box   & 3.820& 0.464  & 12.060    & 0.395     & 4.140 & 0.407 & 2.865 & 0.538         & 3.252    & 0.506            & 4.325  & 0.415& 6.220 & 0.293   & 4.362    & 0.454 & 3.727& 0.485 & \textbf{2.779} \cellcolor[HTML]{EFEFEF} & \textbf{0.540} \cellcolor[HTML]{EFEFEF} \\
guitar       & 1.945& 0.859  & 5.364    & 0.835       & 1.700& 0.861   & 1.172        & 0.931   & 1.046   & 0.941          & 0.883  & 0.944 & 2.044  & 0.823   & 2.242    & 0.855 & 1.022 & 0.915& \textbf{0.622} \cellcolor[HTML]{EFEFEF} & \textbf{0.959} \cellcolor[HTML]{EFEFEF} \\
keyboard     & 2.146 & 0.800 & 2.695     & 0.763      & 2.047  & 0.754& 1.633   & 0.868       & 1.618  & 0.853          & 1.909   & 0.824 & 5.718  & 0.673   & 3.292   & 0.765  & 1.533& 0.836  & \textbf{1.211} \cellcolor[HTML]{EFEFEF} & \textbf{0.892} \cellcolor[HTML]{EFEFEF} \\
lamp         & 13.450& 0.439 & 5.684 & 0.368         & 14.500  & 0.422& 11.892    & 0.613     & 5.368         & 0.701  & 7.568 & 0.680 & 19.102  & 0.395  & 17.918  & 0.460  & 8.069 & 0.501 & \textbf{3.872} \cellcolor[HTML]{EFEFEF} & \textbf{0.709} \cellcolor[HTML]{EFEFEF} \\
laptop       & 4.636 & 0.620 & 5.204    & 0.579       & 4.335& 0.569  & 3.726    & 0.730          & 3.848 & 0.684          & 5.042 & 0.644 & 13.649 & 0.484   & 4.862 & 0.618   & 2.603 & 0.649 & \textbf{2.036} \cellcolor[HTML]{EFEFEF} & \textbf{0.739} \cellcolor[HTML]{EFEFEF} \\
mantel       & 3.525 & 0.496 & 7.120   & 0.468        & 3.752  & 0.455& \textbf{2.339} & \textbf{0.642}  & 2.600     & 0.614         & 2.853 & 0.564   & 6.835   & 0.369  & 4.280 & 0.505   & 3.524 & 0.486& 2.681          \cellcolor[HTML]{EFEFEF} & 0.588          \cellcolor[HTML]{EFEFEF}  \\
monitor      & 3.533 & 0.576 & 9.540   & 0.508         & 3.707& 0.510  & 2.710    & 0.660        & 2.697     & 0.651       & 3.066& 0.582  & 5.600  & 0.453  & 4.584 & 0.546   & 3.510& 0.551 & \textbf{2.676} \cellcolor[HTML]{EFEFEF} & \textbf{0.662} \cellcolor[HTML]{EFEFEF} \\
night\_stand & 4.602& 0.425   & 7.404 & 0.345          & 4.973& 0.378  & 3.603     & 0.511        & \textbf{3.431} & \textbf{0.530}& 4.042 & 0.439 & 8.336    & 0.319& 6.786  & 0.430   & 4.895 & 0.421& 3.728          \cellcolor[HTML]{EFEFEF} & 0.506          \cellcolor[HTML]{EFEFEF} \\
person       & 6.328 & 0.486 & 4.452  & 0.468        & 5.323   & 0.508& 4.452   & 0.629         & 3.502 & 0.690           & 4.025  & 0.651 & 6.579 & 0.517   & 7.651& 0.559     & 4.116 & 0.604& \textbf{2.620} \cellcolor[HTML]{EFEFEF} & \textbf{0.711} \cellcolor[HTML]{EFEFEF} \\
piano        & 6.979  & 0.375   & 6.691  & 0.352        & 6.983  & 0.347& 5.123     & 0.513       & 4.871     & 0.524       & 5.838 & 0.457 & 10.400   & 0.321& 7.312 & 0.412    & 5.364 & 0.427& \textbf{4.056} \cellcolor[HTML]{EFEFEF} & \textbf{0.530} \cellcolor[HTML]{EFEFEF} \\
plant        & 10.652 & 0.278& 16.971       & 0.288   & 9.547 & 0.299 & 7.958 & 0.405          & 6.241  & 0.459          & 7.794  & 0.408 & 10.272 & 0.363  & 10.535  & 0.384 & 8.252 & 0.386 & \textbf{5.562} \cellcolor[HTML]{EFEFEF} & \textbf{0.486} \cellcolor[HTML]{EFEFEF} \\
radio        & 4.859& 0.457  & \textbf{2.640} & 0.354 & 5.268& 0.408  & 3.730 & 0.585           & 3.322   & 0.606           & 3.685  & 0.550  & 9.317  & 0.340  & 8.355    & 0.439& 4.056  & 0.518& 3.158          \cellcolor[HTML]{EFEFEF} & \textbf{0.626} \cellcolor[HTML]{EFEFEF} \\
range\_hood  & 4.447 & 0.445  & 3.612 & 0.429         & 4.858  & 0.384& \textbf{3.274}  & \textbf{0.570}& 3.352      & 0.554            & 3.743  & 0.504 & 7.943 & 0.320    & 5.569& 0.436    & 4.386& 0.421  & 3.388          \cellcolor[HTML]{EFEFEF} & 0.530          \cellcolor[HTML]{EFEFEF} \\
sink         & 5.973& 0.470  & 15.213   & 0.410      & 6.206  & 0.424& 4.267       & 0.638   & 3.843   & 0.647         & 4.428  & 0.617 & 10.124   & 0.383& 7.919   & 0.481  & 5.259& 0.525  & \textbf{3.640} \cellcolor[HTML]{EFEFEF} & \textbf{0.658} \cellcolor[HTML]{EFEFEF} \\
sofa         & 3.453& 0.487  & 12.537 & 0.413        & 3.803 & 0.433 & \textbf{2.832} & \textbf{0.580}& 2.884   & 0.569                 & 3.370 & 0.490   & 6.006 & 0.361   & 4.701  & 0.454    & 3.734 & 0.462& 3.013          \cellcolor[HTML]{EFEFEF} & 0.554          \cellcolor[HTML]{EFEFEF} \\
stairs       & 12.333 & 0.333& 25.836      & 0.312   & 10.243 & 0.377& 8.286   & 0.514          & 5.274  & 0.623        & 6.474  & 0.582& 11.009 & 0.450   & 13.773 & 0.466  & 6.851 & 0.514  & \textbf{4.101} \cellcolor[HTML]{EFEFEF} & \textbf{0.655} \cellcolor[HTML]{EFEFEF} \\
stool        & 8.365  & 0.407& 19.875& 0.368          & 8.213 & 0.418 & 5.431     & 0.621      & 3.752     & \textbf{0.694}      & 4.607  & 0.659& 12.076  & 0.428  & 9.619  & 0.499  & 7.005 & 0.489& \textbf{3.458} \cellcolor[HTML]{EFEFEF} & 0.678          \cellcolor[HTML]{EFEFEF} \\
table        & 3.490  & 0.647  & 9.778    & 0.610       & 3.468   & 0.646& 2.445  & 0.791         & 2.241 & \textbf{0.806}          & 2.733  & 0.759 & 6.815 & 0.537   & 4.688   & 0.657  & 3.160 & 0.676 & \textbf{2.210} \cellcolor[HTML]{EFEFEF} & 0.776          \cellcolor[HTML]{EFEFEF} \\
tent         & 7.210 & 0.393 & 5.122   & 0.316        & 7.913 & 0.341 & 5.440     & 0.549       & 4.972 & 0.565               & 6.207  & 0.498 & 15.554  & 0.313 & 10.084   & 0.413 & 5.493 & 0.458  & \textbf{4.280} \cellcolor[HTML]{EFEFEF} & \textbf{0.597} \cellcolor[HTML]{EFEFEF} \\
toilet       & 4.876& 0.426  & \textbf{2.321}& 0.384  & 5.256  & 0.370& 3.808   & \textbf{0.532}        & 3.775 & 0.517          & 4.445   & 0.452& 8.771 & 0.327    & 6.528  & 0.423  & 5.252 & 0.387& 3.980          \cellcolor[HTML]{EFEFEF} & 0.496          \cellcolor[HTML]{EFEFEF} \\
tv\_stand    & 5.544 & 0.424 & 5.950 & 0.370         & 5.768& 0.384   & 4.608   & 0.509        & 4.100  & 0.526             & 4.946& 0.457  & 8.755  & 0.318  & 7.864 & 0.398   & 5.212 & 0.441 & \textbf{4.031} \cellcolor[HTML]{EFEFEF} & \textbf{0.527} \cellcolor[HTML]{EFEFEF} \\
vase         & 5.750  & 0.471& 5.230   & 0.350        & 5.932  & 0.419 & 4.630    & 0.564       & 4.253     & 0.561      & 5.177     & 0.499  & 11.550  & 0.332  & 8.610   & 0.438  & 5.289 & 0.476& \textbf{4.102} \cellcolor[HTML]{EFEFEF} & \textbf{0.571} \cellcolor[HTML]{EFEFEF}  \\
wardrobe     & 4.375   & 0.532 & 102.146 & 0.395       & 4.694 & 0.447  & 3.602      & \textbf{0.625}    & 3.333   & 0.604        & 3.697  & 0.527   & 8.663& 0.327     & 5.753  & 0.464  & 3.532 & 0.544 & \textbf{3.231} \cellcolor[HTML]{EFEFEF} & 0.611          \cellcolor[HTML]{EFEFEF}  \\
xbox         & 4.573 & 0.499 & 7.406 & 0.421          & 4.763  & 0.431 & 3.540        & 0.616  & 3.285         & 0.567       & 3.802 & 0.502   & 7.443   & 0.320   & 5.623 & 0.476    & 3.453& 0.539 & \textbf{2.801} \cellcolor[HTML]{EFEFEF} & \textbf{0.621} \cellcolor[HTML]{EFEFEF}  \\ \hline
Mean         & 4.852 & 0.526  & 6.615& 0.474           & 4.973  & 0.490 & 3.682       & 0.628    & 3.546      & 0.633      & 3.962   & 0.587 & 36.435 & 0.424   & 31.750   & 0.526& 4.202  & 0.544& \textbf{3.122} \cellcolor[HTML]{EFEFEF} & \textbf{0.641} \cellcolor[HTML]{EFEFEF}  \\ \hline
\end{tabular}
}
\label{tab:mpc2}
\end{table*}

\subsubsection{Fully supervised completion task}
Table~\ref{tab:mpc1} lists the quantitative comparison results to the state-of-the-art methods on the proposed ModelNet-MPC dataset for the fully supervised completion task in terms of mean Chamfer Distance per point (CD, $\times 10^{-3}$) and mean F-Score @ 0.001 (FS).
We can observe that DuInNet achieves the best performance among all methods, and significantly reduces the mean CD by 0.509 ($\downarrow$\;17.3\%) and {increases mean FS by 0.063 ($\uparrow$\;9.8\%), compared with XMFNet~\cite{xmfnet}.
Similarly, XMFNet~\cite{xmfnet} does not effectively utilize the features of images, achieves worse performance than single-modal SOTA methods.
But our DuInNet makes deep interaction between point clouds and images and achieves the best completion performance, even though it has a simpler point cloud generation network than many single-modal methods.
Moreover, Fig.~\ref{fig:modelnet} shows that DuInNet can generate smooth and complete shapes of the airplane and guitar.} 
These comparisons not only demonstrate the effectiveness of DuInNet, but also indicate that the proposed ModelNet-MPC dataset is challenging for current methods, making it a suitable benchmark for the multimodal point cloud completion task.

\begin{table*}[t]
\caption{
Quantitative comparisons to the state-of-the-art methods on the proposed ModelNet-MPC dataset for the \textbf{zero-shot learning} point cloud completion task in terms of Mean Chamfer Distance per point (CD, $\times 10^{-3}$) and Mean F-Score @ 0.001 (FS). Lower CD and higher FS are better. The best results are highlighted in \textbf{boldface}.}
\centering
\renewcommand\arraystretch{1}
\resizebox{\textwidth}{!}{
\begin{tabular}{cc|cc|cc|cc|cc|cc|cc|cc}
\hline   
\multicolumn{2}{c|}{Methods} &
  \multicolumn{6}{c|}{Single-modal } &
  \multicolumn{8}{c}{Multi-modal} \\ \hline
\multicolumn{2}{c|}{\multirow{2}{*}{Categories}} &
  \multicolumn{2}{c|}{FoldingNet~\cite{foldingnet}} &
  \multicolumn{2}{c|}{PCN~\cite{pcn}} &
  \multicolumn{2}{c|}{TopNet~\cite{topnet}} &
  \multicolumn{2}{c|}{ViPC~\cite{vipc}} &
  \multicolumn{2}{c|}{CSDN~\cite{csdn}} &
  \multicolumn{2}{c|}{XMFNet~\cite{xmfnet}} &
  \multicolumn{2}{c}{\cellcolor[HTML]{EFEFEF} \textbf{DuInNet(Ours)}} \\
\multicolumn{2}{c|}{} &
  CD &
  \multicolumn{1}{c|}{FS} &
  CD &
  \multicolumn{1}{c|}{FS} &
  CD &
  \multicolumn{1}{c|}{FS} &
  CD &
  \multicolumn{1}{c|}{FS} &
  CD &
  \multicolumn{1}{c|}{FS} &
  CD &
  \multicolumn{1}{c|}{FS} &
  \cellcolor[HTML]{EFEFEF} \textbf{CD} &
  \cellcolor[HTML]{EFEFEF} \textbf{FS} \\ \hline\hline

\multicolumn{1}{c|}{\multirow{31}{*}{{\rotatebox{90}{Seen object categories}}}} &
  airplane &
  1.573 &
  0.845 &
  1.464 &
  0.854 &
  1.516 &
  0.844 &
  2.483 &
  0.740 &
  1.599 &
  0.797 &
  1.059 &
  0.882 &
  \cellcolor[HTML]{EFEFEF} \textbf{0.755} &
  \cellcolor[HTML]{EFEFEF} \textbf{0.931} \\
\multicolumn{1}{c|}{} &
  bathtub &
  5.050 &
  0.475 &
  5.422 &
  0.434 &
  5.188 &
  0.428 &
  7.668 &
  0.431 &
  5.364 &
  0.514 &
  3.477 &
  0.576 &
  \cellcolor[HTML]{EFEFEF} \textbf{3.262} &
  \cellcolor[HTML]{EFEFEF} \textbf{0.609} \\
\multicolumn{1}{c|}{} &
  bed &
  3.408 &
  0.517 &
  3.690 &
  0.487 &
  3.693 &
  0.477 &
  5.361 &
  0.458 &
  3.953 &
  0.530 &
  2.681 &
  0.589 &
  \cellcolor[HTML]{EFEFEF} \textbf{2.515} &
 \cellcolor[HTML]{EFEFEF} \textbf{0.626} \\
\multicolumn{1}{c|}{} &
  bench &
  4.234 &
  0.607 &
  4.780 &
  0.589 &
  4.509 &
  0.586 &
  7.701 &
  0.560 &
  5.394 &
  0.646 &
  2.568 &
  0.707 &
 \cellcolor[HTML]{EFEFEF} \textbf{2.342} &
 \cellcolor[HTML]{EFEFEF} \textbf{0.746} \\
\multicolumn{1}{c|}{} &
  bookshelf &
  4.279 &
  0.488 &
  4.667 &
  0.461 &
  4.582 &
  0.454 &
  6.059 &
  0.454 &
  4.728 &
  0.523 &
  \textbf{3.105} &
  0.588 &
  \cellcolor[HTML]{EFEFEF} 3.242 &
 \cellcolor[HTML]{EFEFEF} \textbf{0.609} \\
\multicolumn{1}{c|}{} &
  bottle &
  1.903 &
  0.819 &
  2.108 &
  0.778 &
  2.047 &
  0.764 &
  3.185 &
  0.661 &
  2.419 &
  0.752 &
  1.568 &
  0.821 &
  \cellcolor[HTML]{EFEFEF} \textbf{1.393} &
 \cellcolor[HTML]{EFEFEF} \textbf{0.881} \\
\multicolumn{1}{c|}{} &
  car &
  3.707 &
  0.499 &
  3.835 &
  0.476 &
  3.962 &
  0.454 &
  5.336 &
  0.441 &
  4.054 &
  0.507 &
  3.151 &
  0.544 &
  \cellcolor[HTML]{EFEFEF} \textbf{2.830} &
 \cellcolor[HTML]{EFEFEF} \textbf{0.597} \\
\multicolumn{1}{c|}{} &
  chair &
  4.235 &
  0.551 &
  4.565 &
  0.552 &
  4.228 &
  0.535 &
  7.920 &
  0.539 &
  4.884 &
  0.609 &
  3.026 &
  0.658 &
 \cellcolor[HTML]{EFEFEF} \textbf{2.743} &
 \cellcolor[HTML]{EFEFEF} \textbf{0.692} \\
\multicolumn{1}{c|}{} &
  cone &
  2.910 &
  0.738 &
  3.084 &
  0.684 &
  3.111 &
  0.664 &
  5.244 &
  0.577 &
  4.062 &
  0.693 &
  2.402 &
  0.776 &
 \cellcolor[HTML]{EFEFEF} \textbf{1.921} &
  \cellcolor[HTML]{EFEFEF} \textbf{0.836} \\
\multicolumn{1}{c|}{} &
  desk &
  5.967 &
  0.461 &
  6.735 &
  0.430 &
  6.048 &
  0.449 &
  9.381 &
  0.464 &
  6.736 &
  0.533 &
  3.982 &
  0.596 &
 \cellcolor[HTML]{EFEFEF} \textbf{3.938} &
 \cellcolor[HTML]{EFEFEF} \textbf{0.614} \\
\multicolumn{1}{c|}{} &
  door &
  1.962 &
  0.812 &
  2.035 &
  0.821 &
  2.225 &
  0.790 &
  3.329 &
  0.710 &
  2.260 &
  0.766 &
  1.082 &
  0.837 &
  \cellcolor[HTML]{EFEFEF} \textbf{1.041} &
 \cellcolor[HTML]{EFEFEF} \textbf{0.890} \\
\multicolumn{1}{c|}{} &
  dresser &
  3.489 &
  0.515 &
  3.670 &
  0.456 &
  3.754 &
  0.444 &
  5.667 &
  0.408 &
  4.702 &
  0.484 &
  3.060 &
  0.580 &
 \cellcolor[HTML]{EFEFEF} \textbf{2.615} &
  \cellcolor[HTML]{EFEFEF} \textbf{0.616} \\
\multicolumn{1}{c|}{} &
  flower\_pot &
  8.873 &
  0.304 &
  8.638 &
  0.314 &
  8.075 &
  0.306 &
  9.366 &
  0.385 &
  7.681 &
  0.414 &
  5.148 &
  0.477 &
 \cellcolor[HTML]{EFEFEF} \textbf{4.918} &
 \cellcolor[HTML]{EFEFEF} \textbf{0.492} \\
\multicolumn{1}{c|}{} &
  glass\_box &
  3.533 &
  0.504 &
  3.655 &
  0.445 &
  3.605 &
  0.449 &
  4.985 &
  0.410 &
  4.410 &
  0.482 &
  2.574 &
  0.560 &
 \cellcolor[HTML]{EFEFEF} \textbf{2.535} &
 \cellcolor[HTML]{EFEFEF} \textbf{0.606} \\
\multicolumn{1}{c|}{} &
  guitar &
  1.240 &
  0.901 &
  1.156 &
  0.909 &
  1.084 &
  0.901 &
  1.453 &
  0.857 &
  1.159 &
  0.869 &
  0.488 &
  0.956 &
 \cellcolor[HTML]{EFEFEF} \textbf{0.382} &
 \cellcolor[HTML]{EFEFEF} \textbf{0.974} \\
\multicolumn{1}{c|}{} &
  lamp &
  11.224 &
  0.468 &
  17.241 &
  0.444 &
  14.790 &
  0.456 &
  20.275 &
  0.484 &
  11.078 &
  0.597 &
  5.102 &
  0.667 &
 \cellcolor[HTML]{EFEFEF} \textbf{5.016} &
 \cellcolor[HTML]{EFEFEF}\cellcolor[HTML]{EFEFEF} \textbf{0.685} \\
\multicolumn{1}{c|}{} &
  laptop &
  2.621 &
  0.707 &
  2.638 &
  0.671 &
  2.869 &
  0.619 &
  5.338 &
  0.576 &
  3.091 &
  0.661 &
  1.556 &
  0.719 &
 \cellcolor[HTML]{EFEFEF} \textbf{1.249} &
 \cellcolor[HTML]{EFEFEF}\cellcolor[HTML]{EFEFEF} \textbf{0.804} \\
\multicolumn{1}{c|}{} &
  mantel &
  3.085 &
  0.542 &
  3.208 &
  0.512 &
  3.322 &
  0.495 &
  6.223 &
  0.451 &
  3.997 &
  0.545 &
  2.345 &
  0.648 &
 \cellcolor[HTML]{EFEFEF} \textbf{2.196} &
  \cellcolor[HTML]{EFEFEF}\cellcolor[HTML]{EFEFEF} \textbf{0.655} \\
\multicolumn{1}{c|}{} &
  monitor &
  3.274 &
  0.604 &
  3.465 &
  0.570 &
  3.553 &
  0.538 &
  5.134 &
  0.514 &
  3.847 &
  0.575 &
  2.462 &
  0.636 &
 \cellcolor[HTML]{EFEFEF} \textbf{2.389} &
 \cellcolor[HTML]{EFEFEF} \textbf{0.682} \\
\multicolumn{1}{c|}{} &
  night\_stand &
  4.525 &
  0.436 &
  4.835 &
  0.410 &
  4.843 &
  0.400 &
  7.378 &
  0.406 &
  5.881 &
  0.462 &
  3.712 &
  0.551 &
 \cellcolor[HTML]{EFEFEF} \textbf{3.506} &
\cellcolor[HTML]{EFEFEF}  \textbf{0.563} \\
\multicolumn{1}{c|}{} &
  person &
  6.025 &
  0.468 &
  6.375 &
  0.516 &
  5.480 &
  0.509 &
  6.045 &
  0.589 &
  5.312 &
  0.645 &
  2.451 &
  0.706 &
\cellcolor[HTML]{EFEFEF}  \textbf{2.235} &
\cellcolor[HTML]{EFEFEF}  \textbf{0.733} \\
\multicolumn{1}{c|}{} &
  piano &
  6.264 &
  0.405 &
  6.393 &
  0.385 &
  6.096 &
  0.380 &
  7.709 &
  0.411 &
  5.756 &
  0.480 &
  3.700 &
  0.548 &
\cellcolor[HTML]{EFEFEF}  \textbf{3.449} &
 \cellcolor[HTML]{EFEFEF} \textbf{0.583} \\
\multicolumn{1}{c|}{} &
  plant &
  11.297 &
  0.249 &
  11.580 &
  0.292 &
  9.598 &
  0.305 &
  9.507 &
  0.436 &
  8.376 &
  0.459 &
  5.669 &
  \textbf{0.517} &
 \cellcolor[HTML]{EFEFEF} \textbf{5.414} &
\cellcolor[HTML]{EFEFEF}  0.515 \\
\multicolumn{1}{c|}{} &
  range\_hood &
  4.086 &
  0.485 &
  4.470 &
  0.443 &
  4.432 &
  0.417 &
  7.587 &
  0.408 &
  4.889 &
  0.498 &
  3.152 &
  0.580 &
\cellcolor[HTML]{EFEFEF}  \textbf{2.905} &
\cellcolor[HTML]{EFEFEF}  \textbf{0.587} \\
\multicolumn{1}{c|}{} &
  sofa &
  3.299 &
  0.514 &
  3.533 &
  0.482 &
  3.530 &
  0.475 &
  5.337 &
  0.435 &
  4.050 &
  0.495 &
  2.834 &
  0.552 &
\cellcolor[HTML]{EFEFEF}  \textbf{2.605} &
 \cellcolor[HTML]{EFEFEF} \textbf{0.599} \\
\multicolumn{1}{c|}{} &
  table &
  3.287 &
  0.683 &
  4.017 &
  0.652 &
  3.262 &
  0.679 &
  6.199 &
  0.606 &
  3.745 &
  0.691 &
  2.115 &
  0.751 &
 \cellcolor[HTML]{EFEFEF} \textbf{1.968} &
\cellcolor[HTML]{EFEFEF}  \textbf{0.800} \\
\multicolumn{1}{c|}{} &
  toilet &
  4.551 &
  0.457 &
  4.983 &
  0.412 &
  4.935 &
  0.399 &
  8.034 &
  0.407 &
  5.699 &
  0.461 &
  3.791 &
  0.517 &
  \cellcolor[HTML]{EFEFEF} \textbf{3.579} &
 \cellcolor[HTML]{EFEFEF} \textbf{0.545} \\
\multicolumn{1}{c|}{} &
  tv\_stand &
  5.138 &
  0.461 &
  5.744 &
  0.408 &
  5.545 &
  0.419 &
  7.236 &
  0.410 &
  5.975 &
  0.477 &
  3.735 &
  0.550 &
\cellcolor[HTML]{EFEFEF}  \textbf{3.727} &
\cellcolor[HTML]{EFEFEF}  \textbf{0.575} \\
\multicolumn{1}{c|}{} &
  vase &
  5.054 &
  0.510 &
  5.455 &
  0.461 &
  5.351 &
  0.454 &
  8.597 &
  0.444 &
  7.132 &
  0.507 &
  4.019 &
  0.580 &
 \cellcolor[HTML]{EFEFEF} \textbf{3.667} &
\cellcolor[HTML]{EFEFEF}  \textbf{0.620} \\
\multicolumn{1}{c|}{} &
  xbox &
  3.811 &
  0.559 &
  4.479 &
  0.499 &
  4.464 &
  0.474 &
  5.830 &
  0.439 &
  4.098 &
  0.546 &
  2.433 &
  0.631 &
 \cellcolor[HTML]{EFEFEF} \textbf{2.270} &
 \cellcolor[HTML]{EFEFEF} \textbf{0.693} \\ \cline{2-16} 
\multicolumn{1}{c|}{} &
  Mean(seen) &
  4.132 &
  0.559 &
  4.487 &
  0.535 &
  4.275 &
  0.526 &
  6.375 &
  0.505 &
  4.646 &
  0.573 &
  2.894 &
  0.641 &
\cellcolor[HTML]{EFEFEF}  \textbf{2.703} &
 \cellcolor[HTML]{EFEFEF} \textbf{0.676} \\ \hline
\multicolumn{1}{c|}{\multirow{11}{*}{{\rotatebox{90}{Unseen object categories}}}} &
  bowl &
  12.270 &
  0.280 &
  13.563 &
  0.225 &
  11.459 &
  0.229 &
  24.103 &
  0.337 &
  19.152 &
  0.374 &
  8.381 &
\textbf{0.477} &
 \cellcolor[HTML]{EFEFEF} \textbf{7.006} &
  \cellcolor[HTML]{EFEFEF}  0.456 \\
\multicolumn{1}{c|}{} &
  cup &
  8.726 &
  0.327 &
  8.951 &
  0.285 &
  8.865 &
  0.283 &
  11.192 &
  0.365 &
  9.300 &
  0.380 &
  6.017 &
  0.472 &
\cellcolor[HTML]{EFEFEF}  \textbf{5.531} &
\cellcolor[HTML]{EFEFEF}  \textbf{0.487} \\
\multicolumn{1}{c|}{} &
  curtain &
  3.054 &
  0.652 &
  3.311 &
  0.674 &
  3.165 &
  0.676 &
  6.679 &
  0.616 &
  5.210 &
  0.663 &
  1.596 &
  0.803 &
\cellcolor[HTML]{EFEFEF}  \textbf{1.336} &
\cellcolor[HTML]{EFEFEF}  \textbf{0.839} \\
\multicolumn{1}{c|}{} &
  keyboard &
  2.163 &
  0.717 &
  3.137 &
  0.710 &
  2.340 &
  0.669 &
  3.560 &
  0.693 &
  4.831 &
  0.653 &
  1.091 &
  0.810 &
 \cellcolor[HTML]{EFEFEF} \textbf{0.929} &
 \cellcolor[HTML]{EFEFEF} \textbf{0.885} \\
\multicolumn{1}{c|}{} &
  radio &
  6.472 &
  0.387 &
  6.587 &
  0.363 &
  6.335 &
  0.361 &
  8.377 &
  0.422 &
  6.993 &
  0.468 &
  3.254 &
  0.597 &
\cellcolor[HTML]{EFEFEF}  \textbf{3.026} &
\cellcolor[HTML]{EFEFEF}  \textbf{0.622} \\
\multicolumn{1}{c|}{} &
  sink &
  10.555 &
  0.353 &
  11.331 &
  0.317 &
  10.455 &
  0.324 &
  11.516 &
  0.430 &
  8.634 &
  0.486 &
  \textbf{4.842} &
  0.578 &
 \cellcolor[HTML]{EFEFEF} 4.884 &
\cellcolor[HTML]{EFEFEF}  \textbf{0.587} \\
\multicolumn{1}{c|}{} &
  stairs &
  21.967 &
  0.215 &
  18.199 &
  0.269 &
  16.866 &
  0.264 &
  12.988 &
  0.484 &
  12.760 &
  0.507 &
  \textbf{5.572} &
  \textbf{0.599} &
\cellcolor[HTML]{EFEFEF}  5.713 &
\cellcolor[HTML]{EFEFEF}  0.587 \\
\multicolumn{1}{c|}{} &
  stool &
  13.104 &
  0.375 &
  10.893 &
  0.398 &
  9.642 &
  0.383 &
  12.930 &
  0.487 &
  8.297 &
  0.532 &
  4.802 &
  0.625 &
 \cellcolor[HTML]{EFEFEF} \textbf{4.488} &
 \cellcolor[HTML]{EFEFEF} \textbf{0.635} \\
\multicolumn{1}{c|}{} &
  tent &
  9.371 &
  0.333 &
  9.721 &
  0.295 &
  9.701 &
  0.298 &
  12.816 &
  0.409 &
  10.639 &
  0.465 &
  5.134 &
  0.561 &
 \cellcolor[HTML]{EFEFEF} \textbf{4.918} &
 \cellcolor[HTML]{EFEFEF} \textbf{0.563} \\
\multicolumn{1}{c|}{} &
  wardrobe &
  3.868 &
  0.515 &
  4.568 &
  0.429 &
  4.560 &
  0.432 &
  5.833 &
  0.426 &
  5.843 &
  0.454 &
  2.807 &
  0.607 &
 \cellcolor[HTML]{EFEFEF} \textbf{2.536} &
 \cellcolor[HTML]{EFEFEF} \textbf{0.654} \\ \cline{2-16} 
\multicolumn{1}{c|}{} &
  Mean(unseen) &
  8.950 &
  0.428 &
  8.810 &
  0.412 &
  8.178 &
  0.407 &
  10.424 &
  0.481 &
  8.787 &
  0.512 &
  4.102 &
  0.627 & 
\cellcolor[HTML]{EFEFEF}  \textbf{3.847} &
\cellcolor[HTML]{EFEFEF}  \textbf{0.648} \\ \hline
 \multicolumn{1}{c|}{} &
  Mean &
  5.066 &
  0.534 &
  5.325 &
  0.511 &
  5.032 &
  0.503 &
  7.160 &
  0.501 &
  5.449 &
  0.562 &
  3.128 &
  0.638 &
 \cellcolor[HTML]{EFEFEF} \textbf{2.925} &
\cellcolor[HTML]{EFEFEF}  \textbf{0.671} \\ \hline
\end{tabular}
}
\label{tab:mpc3}
\end{table*}

\subsubsection{Denoising completion task}
The noisy partial point cloud challenges the shape completion task due to its perturbations of shape prior information. To evaluate the effectiveness of the proposed method on the denoising completion task, we conduct experiments on the proposed ModelNet-MPC dataset.
Table~\ref{tab:mpc2} lists the quantitative comparison results to the state-of-the-art methods on the proposed ModelNet-MPC dataset for the denoising completion task in terms of mean CD and FS.
It is worth noting that, due to the excessive reliance on the input point clouds, ViPC~\cite{vipc} and CSDN~\cite{csdn} fail to restore complete point clouds in some specific categories (have high Chamfer Distances) when the input point clouds are noisy.
It proves that these two methods have poor noise robustness.
But based on the feature interaction and shape prior exploration in these two modalities, DuInNet still successfully restores the complete point clouds, and reduces mean CD by 1.080 ($\downarrow$\;25.7\%) and {increases mean FS by 0.097 ($\uparrow$\;17.8\%), compared with XMFNet~\cite{xmfnet}.
Moreover, Fig.~\ref{fig:modelnet} shows that DuInNet can still complete the laptop monitor in the presence of noise, thanks to the assistance of the image path, but XMFNet~\cite{xmfnet} fails in this situation.
Above experiments demonstrate that DuInNet has good noise robustness for the point cloud completion task, which is important for real-world scan applications.}

\subsubsection{Zero-shot learning completion task}
To evaluate the transfer ability of DuInNet on unseen categories, we conduct experiments on the ModelNet-MPC dataset. Specifically, 30 categories are used to train the network, and 40 categories (including 10 unseen categories) are used to test the zero-shot transfer ability. 
Table~\ref{tab:mpc3} lists the quantitative comparison results to the state-of-the-art methods on the proposed ModelNet-MPC dataset for the zero-shot learning completion task in terms of mean CD and FS.
We can observe that DuInNet still achieves the best performance among all methods and in all terms including seen, unseen and overall scenarios. 
In particular, DuInNet can explore more precise shape-specific descriptors by leveraging multimodal feature interaction. When dealing with unseen object categories, DuInNet can still perform the completion task satisfactorily. 

In summary, the above extensive quantitative and qualitative experiments demonstrate that DuInNet outperforms the state-of-the-art methods on the proposed ModelNet-MPC dataset, and has good noise robustness, category generalization and zero-shot transfer ability.

\subsection{Ablation Studies}
In this section, to evaluate the effectiveness of contributions of the dual feature interactor and the adaptive point generator, we conduct ablation experiments with different input modalities, fusion strategies, and partition of image and point cloud blocks on the ModelNet-MPC dataset.
Specifically, as shown in Table~\ref{tab:ablation1}, model A and B only use images or point clouds as the single-modal input. 
Model C and D respectively take one input modality (images or point clouds) as query tensors to fuse features with the other modality, which stand for the previous unidirectional fusion mode.
Model E is our proposed DuInNet that adopts a dual-path feature interaction architecture.
As shown in Table~\ref{tab:ablation2}, model F-J represent different partitions of image and point cloud blocks of the adaptive point generator for point cloud generations.

\begin{table}[t]
\centering
\renewcommand\arraystretch{1.1}
\caption{Ablation studies of DFI about different input modalities and fusion strategies on the ModelNet-MPC dataset.}
\resizebox{0.34\textwidth}{!}{
\begin{tabular}{c|cc|cc}\hline
Model       & Images           & Points           & CD             & FS            \\\hline \hline
A          & \ding{52}              &   \ding{55}                & 5.342          & 0.593          \\
B          &      \ding{55}             & \ding{52}               & 2.954          & 0.662          \\
C          & \multicolumn{2}{c|}{\ding{52}\;\;$\Longleftarrow$\quad\ding{52}}        & 2.566          & 0.687          \\
D          & \multicolumn{2}{c|}{\ding{52}\;\;$\Longrightarrow$\quad\ding{52}}         & 2.712          & 0.670          \\\hline
\rowcolor[HTML]{EFEFEF} \textbf{E} & \multicolumn{2}{c|}{\ding{52}\;\;$\Longleftrightarrow$\;\;\;\ding{52}} & \textbf{2.428} & \textbf{0.707}\\\hline
\end{tabular}
\label{tab:ablation1}}
\end{table}
\begin{table}[t]
\centering
\renewcommand\arraystretch{1.2}
\caption{Ablation studies of APG about different partitions of image and point cloud blocks on the ModelNet-MPC dataset.}
\resizebox{0.48\textwidth}{!}{
\begin{tabular}{c|cc|cc|cc|cc}
\hline
\multirow{2}{*}{Model} & \multirow{2}{*}{$N_{img}$} & \multirow{2}{*}{$N_{pc}$} & \multicolumn{2}{c|}{Fully supervision} & \multicolumn{2}{c|}{Denoising} & \multicolumn{2}{c}{Zero-shot learning} \\
  &    &    & CD             & FS             & CD             & FS            & CD             & FS             \\ \hline \hline
F & 0  & 16 & 2.637          & 0.668          & \textbf{3.122} & \textbf{0.641} & 3.112          & 0.645          \\
G & 4  & 12 & 2.617          & 0.659          & 3.198          & 0.628         & 3.013          & 0.654          \\
H & 8  & 8  & 2.585          & 0.669          & 3.269          & 0.613         & \textbf{2.925} & \textbf{0.671} \\
I & 12 & 4  & 2.588          & 0.680           & 3.404          & 0.609         & 2.943          & 0.664          \\
J & 16 & 0  & \textbf{2.428} & \textbf{0.707} & 3.437          & 0.605         & 2.955          & 0.662          \\ \hline
\end{tabular}
}
\label{tab:ablation2}
\end{table}
From Table~\ref{tab:ablation1}, we can observe that, compared with models A and B, the multimodal methods (models C, D and E) utilize more shape prior information, and thus obtain better point cloud completion performance.
{
As for the single-modal point cloud completion task, point clouds are more suitable to generate completion shapes than images.}
Meanwhile, the comparative results between the last three multimodal models demonstrate that the dual-path feature interaction strategy of DFI achieves better performance than the single-path unidirectional fusion strategy. 

From Table~\ref{tab:ablation2}, we can observe that APG can set different optimal partitions for different tasks, which is also crucial for real-world scan applications. 
Moreover, Fig.~\ref{fig:modelnet} shows that DuInNet can adaptively generate high-quality complete point clouds in blocks with different weights for two modalities according to the specific requirements of tasks.
Specifically, the full supervision and denoising completion tasks separately prefer more image and point cloud blocks, due to the need for clearer object boundaries of images and stronger point cloud feature discriminative ability, respectively.
As for the zero-shot learning task, APG needs to simultaneously consider two modalities to deal with unseen object categories with a better transfer ability. 

In summary, these experiments demonstrate the effectiveness of DFI and APG.
\section{Conclusion}
\label{sec:conclusion}
In this paper, we propose a dual modality feature interaction network DuInNet for point cloud completion. Both geometric and texture characteristics of shapes are explored using the dual feature interactor, and complete point clouds are adaptively generated in blocks with different weights for two modalities by the adaptive point generator.
Meanwhile, to further boost the development of multimodal point cloud completion, we introduce a new multimodal point cloud completion benchmark ModelNet-MPC, which contains nearly 400,000 pairs of high-quality point clouds and {rendered} images for 40 categories. Two additional completion tasks including denoising completion and zero-shot learning completion are proposed to evaluate the robustness and transfer ability of methods.
Extensive experiments on the ShapeNet-ViPC and ModelNet-MPC datasets demonstrate the superiority, noise robustness, category generalization and zero-shot transfer ability of our DuInNet.

\normalem
\bibliography{Main}
\bibliographystyle{IEEEtran}

\end{document}